\documentclass{article}

\usepackage{microtype}
\usepackage{graphicx}
\usepackage{subcaption}
\usepackage{booktabs} 

\usepackage{hyperref}

\usepackage[accepted]{icml2026}

\usepackage{amsmath}
\usepackage{amssymb}
\usepackage{mathtools}
\usepackage{amsthm}

\usepackage[capitalize,noabbrev]{cleveref}

\usepackage{longtable}

\usepackage{graphicx}
\usepackage{subcaption} 
\usepackage{hyperref}
\usepackage{url}
\usepackage{algorithm}

\usepackage{booktabs,  amsmath}
\usepackage[normalem]{ulem}
\usepackage{multirow}
\usepackage{graphicx}
\usepackage{xcolor}
\usepackage{colortbl}
\usepackage{booktabs}
\usepackage{threeparttable}
\usepackage{siunitx} 
\usepackage{amsmath} 
\usepackage{makecell, booktabs}
\usepackage{wrapfig}
\theoremstyle{plain}

\theoremstyle{definition}

\theoremstyle{remark}

\usepackage[textsize=tiny]{todonotes}

\icmltitlerunning{Adaptive Testing for LLM Evaluation: A Psychometric Alternative to Static Benchmarks}
\begin{document}

\twocolumn[
  \icmltitle{Adaptive Testing for LLM Evaluation:\\ A Psychometric Alternative to Static Benchmarks}



  \icmlsetsymbol{equal}{*}

  \begin{icmlauthorlist}
    \icmlauthor{Peiyu Li}{equal,yyy}
    \icmlauthor{Xiuxiu Tang}{equal,yyy}
    \icmlauthor{Si Chen}{yyy}
    \icmlauthor{Ying Cheng}{yyy}
    \icmlauthor{Ronald Metoyer}{yyy}
    \icmlauthor{Ting Hua}{yyy}
    \icmlauthor{Nitesh V Chawla}{yyy}
  \end{icmlauthorlist}

  \icmlaffiliation{yyy}{University of Notre Dame, Notre Dame, USA}

\icmlcorrespondingauthor{Ying Cheng}{ycheng4@nd.edu}
  \icmlcorrespondingauthor{Ting Hua}{thua@nd.edu}
  \icmlcorrespondingauthor{Nitesh V Chawla}{nchawla@nd.edu}

  \icmlkeywords{Machine Learning, ICML}

  \vskip 0.3in

]



\printAffiliationsAndNotice{\icmlEqualContribution}
\newcommand{\model}{ATLAS}
\newcommand{\CAT}{ATLAS}

\newcommand{\ting}[1]{{\bf \color{cyan}{[Ting: #1]}}}
\newcommand{\xiuxiu}[1]{{\bf \color{red}{[Xiuxiu: #1]}}}
\newcommand{\peiyu}[1]{{\bf \color{blue}{[Georgia: #1]}}}
\begin{abstract}
Evaluating large language models (LLMs) typically requires thousands of benchmark items, making the process expensive, slow, and increasingly impractical at scale. Existing evaluation protocols rely on average accuracy over fixed item sets, treating all items as equally informative despite substantial variation in difficulty and discrimination. We introduce \CAT, an adaptive testing framework based on Item Response Theory (IRT) that estimates model ability using Fisher information–guided item selection. \CAT\ reduces the number of required items by up to 90\% while maintaining measurement precision. For instance, it matches whole-bank ability estimates using only 41 items (0.157 MAE) on HellaSwag (5,600 items). We further reconstruct accuracy from \CAT's ability estimates and find that reconstructed accuracies closely match raw accuracies across all five benchmarks, indicating that ability $\theta$ preserves the global performance structure. At the same time, $\theta$ provides finer discrimination within accuracy-equivalent models: among more than 3{,}000 evaluated models, 23--31\% shift by more than 10 rank positions, and models with identical accuracies receive meaningfully different ability estimates. Code and calibrated item banks available at \url{https://github.com/Peiyu-Georgia-Li/ATLAS}. 
\end{abstract}

\section{Introduction}

Large language model evaluation relies on benchmarks with tens of thousands of items, which are costly to run and often take days or weeks to complete. Even with benchmarks exceeding 100,000 items, evaluation still depends on average accuracy over fixed item sets. This practice overlooks valuable statistical information and raises concerns about efficiency and validity.

Current evaluation practices face three fundamental limitations. First, average accuracy over fixed item sets obscure meaningful differences between models with distinct error patterns, especially
among lower-performing models where small ability differences are dominated by measurement
noise. Second, static evaluations treat poorly discriminative items as equally informative as
high-quality questions, leading to unreliable and often misleading comparisons. Third, evaluating
complete benchmarks is inefficient and time-consuming, requiring models to answer hundreds
or thousands of items regardless of how much additional information those items provide.

To address these limitations, we propose \CAT\ (Adaptive Testing for LLM Ability Scoring), an adaptive evaluation framework based on computerized adaptive testing (CAT) \citep{lord1980, wainer2000computerized, weiss1982improving}. \CAT\ first calibrates benchmark items using three-parameter logistic (3PL) IRT models to estimate item difficulty, discrimination, and guessing parameters \citep{birnbaum1968some,hambleton1991fundamentals}. Then, rather than administering fixed item sets, \CAT\ dynamically selects items with maximum Fisher information for each model's current  estimated ability, terminating when precision thresholds are reached. This approach directly addresses all three limitations: Fisher information-guided selection provides precise ability estimates that distinguish models with identical accuracy, dynamic item selection
prioritizes highly discriminative items rather than treating all questions as
equally informative, and adaptive termination enables reliable evaluation with far fewer items
and substantially less time than full-benchmark scoring.

We evaluate \CAT\ across five major benchmarks, including WinoGrande \citep{ai2:winogrande},
TruthfulQA \citep{lin2021truthfulqa}, HellaSwag \citep{zellers2019hellaswag}, GSM8K
\citep{cobbe2021gsm8k}, ARC \citep{allenai:arc} and find that it matches or exceeds the
accuracy of strong static baselines while using far fewer items. For example, \CAT\
achieves the lowest MAE on TruthfulQA ($0.064$ with 48 items) and HellaSwag
($0.157$ with 41 items), and matches MetaBench \citep{metabench} on WinoGrande
while using 2× fewer items (70 vs.\ 133). It also outperforms TinyBenchmarks
\citep{polo2024tinybenchmarks}, which uses 97–100 items but yields higher error across all
benchmarks. Overall, \CAT\ requires only 30–89 items per benchmark compared to
hundreds in static subsets, and attains the lowest Information Efficiency Score (IES)
across all benchmarks, demonstrating the strongest accuracy–efficiency tradeoff.


Our contributions are: 
(1) We identify fundamental limitations of raw accuracy-based evaluation and show that IRT-based ability estimates provide more informative comparisons of LLM performance. In particular, ability-based evaluation reveals substantial rank reordering and produces more stable model rankings across benchmark subsets and related domains.
(2) We introduce \CAT, a large-scale adaptive testing framework for LLMs that achieves up to 90\% item reduction while maintaining target measurement precision through SE-controlled stopping, enabling scalable and precision-targeted evaluation beyond fixed-length adaptive designs.
(3) We evaluate the robustness of ATLAS beyond the calibration distribution through strict architecture-family holdout and temporal generalization experiments, showing that calibrated item parameters transfer reasonably well to unseen model families and newer model populations.
(4) We emphasize rigorous psychometric validation by reporting model-fit diagnostics and demonstrating the use of common-person linking to efficiently align item parameters and ensure scalable cross-model comparability.
\section{Related Work}

\subsection{IRT-Based Approaches}
Item Response Theory (IRT) has recently been applied to LLM evaluation \citep{lalor2024item, habba2026growing}. It provides item parameters such as difficulty, discrimination, and guessing, as well as latent ability estimates $\theta$ for models. However, existing IRT applications remain largely \textbf{static} in nature. For instance, TinyBenchmarks \citep{polo2024tinybenchmarks} uses clustering for item selection but doesn't guarantee informativeness for $\theta$ estimation, while MetaBench \citep{metabench} requires computationally expensive iterations to identify stable subsets. Moreover, these approaches often lack proper psychometric validation and emphasize predictive accuracy over \textbf{model fit}. TinyBenchmarks and MetaBench do not report fit statistics. Instead, we compute
these metrics using their released IRT code (as shown in Table~\ref{m2_compare}). This limitation makes it difficult to ensure that the resulting ability estimates are valid, interpretable, and comparable across models.
A detailed comparison of IRT-based approaches is provided in Appendix~\ref{app:method_comparison}.

Beyond these limitations of existing IRT applications, many evaluations continue to rely on average scores. Average scores tend to mask meaningful model differences and are often affected by form-dependence, nonlinear scaling, equal weighting of uninformative items, and contamination sensitivity (see Appendix~\ref{app:avg_score_limits} for detailed analysis). In contrast, IRT-based ability estimates ($\theta$) provide form-invariant, uncertainty-aware alternatives that adjust for item difficulty and discrimination.

\subsection{Adaptive Testing} 


Computerized adaptive testing (CAT) adjusts item administration based on an examinee’s evolving ability estimate \citep{meijer1999computerized, van2010elements}. After each response, the test updates the ability estimate and selects the next item using an algorithm that aims to provide the most informative measurement while satisfying test constraints \citep{weiss1982improving, chang2015psychometrics, cheng2009maximum}. Related adaptive frameworks such as multistage testing and process-data–based approaches apply similar principles and offer additional flexibility and diagnostic information \citep{zenisky2009multistage, zheng2015fly, tang2024utilizing}. These features allow CAT to evaluate examinees efficiently while maintaining rigorous measurement precision. This adaptive structure aligns well with challenges in evaluating LLMs, which vary widely in their performance levels. Current evaluations often use large static benchmarks in which every model must answer all items, even when many items provide little information about its ability. These benchmarks also rarely report empirical item characteristics, so their difficulty range and informativeness across models remain unclear. CAT addresses these limitations by selecting items targeted to each model’s estimated ability, which yields more precise and efficient evaluation with far fewer items.

However, only a few studies have explored adaptive evaluation for LLMs. Early efforts were either limited in scope or primarily conceptual \citep{zhuangposition}. A recent study that is closely related to our work is Fluid Benchmarking \citep{hofmann2025fluid}, which appeared concurrently with this research. Fluid also applies CAT principles to increase evaluation efficiency and models LLM performance on a latent ability scale. The two approaches are complementary and differ in several ways. Fluid focuses on adaptive evaluation during LLM pretraining, whereas our work examines post-training evaluation. Fluid calibrates its IRT model on 102 LMs, while ATLAS uses a  substantially larger and more diverse pool of 3{,}000+ LMs. Fluid adopts a fixed-length adaptive design, while ATLAS uses a precision-based stopping rule that terminates the test once the uncertainty of the abilityestimate falls below a predefined threshold. Precision-based stopping ensures consistent measurement precision across models and avoids administering unnecessary items. In addition, we report model-fit statistics to ensure the adequacy of the IRT model before running CAT and provide a transparent description of calibration and linking procedures used to estimate item parameters. Both studies demonstrate that adaptive testing methods can be used at different stages of LLM development and under different design choices, which illustrates the broader potential of CAT-based approaches for scalable and precise LLM assessment.

\section{Methodology}

We introduce a novel adaptive testing framework that transforms LLM evaluation from static benchmarking to dynamic ability estimation. Our approach addresses three critical limitations of current evaluation practice:  (1) it reduces computational cost by requiring 90\% fewer items while maintaining overall performance, (2) it overcomes the
ceiling effects of accuracy-based metrics and preserves discrimination across the
ability spectrum, and (3) it distinguishes models with identical accuracy score but
different abilities.

This section presents our framework in four stages: problem formulation (Section~\ref{sec:problem_formulation}), data construction with psychometric filtering (Section~\ref{sec:data_construction}), item bank calibration using IRT models (Section~\ref{sec:calibration}), and adaptive testing with randomesque selection (Section~\ref{sec:adaptive_testing}).

\subsection{Problem Formulation and Setup}\label{sec:problem_formulation}

We formulate LLM evaluation as a psychometric measurement problem. Let $\mathcal{I}$ denote the set of benchmark items and $\mathcal{L}$ the set of language models. For each model $\ell \in \mathcal{L}$ and item $i \in \mathcal{I}$, we observe a binary response $Y_{i,\ell} \in \{0,1\}$, where 1 indicates correct and 0 incorrect. These responses form the item-response matrix $\{Y_{i,\ell}\}_{i \in \mathcal{I}, \ell \in \mathcal{L}}$.

Unlike traditional approaches that rely solely on accuracy scores, our objective is to estimate the latent ability $\theta_\ell$ of each model based on its response pattern $\{Y_{i,\ell}\}_{i \in \mathcal{I}}$, while simultaneously calibrating item-level parameters: discrimination $a_i$, difficulty $b_i$, and lower-asymptote $c_i$. This approach enables fine-grained model comparison even when models achieve identical accuracy, as $\theta_\ell$ accounts for the varying informativeness of different items.

\subsection{Data Construction with Psychometric Filtering}\label{sec:data_construction}

We construct the item-response matrix using data from the HuggingFace Open LLM Leaderboard. The item pool $\mathcal{I}$ spans five benchmarks: ARC, GSM8K, HellaSwag, TruthfulQA, and WinoGrande. To ensure data quality for IRT calibration, we apply two levels of filtering: removing unsuitable models and eliminating non-informative items.

\paragraph{Model Selection and Splitting.} We retain only models $\mathcal{L}$ with complete responses across all items. To obtain a
calibration sample whose ability distribution approximates a Gaussian, a standard assumption for
stable IRT estimation, we exclude models in the extreme low-ability tail (below the 0.1st percentile),
whose near-zero response patterns destabilize 3PL parameter estimation. The high-ability tail is
small and non-degenerate, so these models are retained. The selected models
are then split into training and testing sets using stratified random sampling (10
bins) to ensure that both splits share a similar ability distribution. We allocate 90\% of the
models to the training set for item calibration, and use the remaining 10\% as the
testing set for evaluating performance in our
experiments (see Table~\ref{tab:filtering_summary}).

\paragraph{Item Filtering.} We apply two complementary filters to retain only discriminative items:
\begin{itemize}
    \item \textbf{Low-variance removal:} Items with response standard deviation $< 1\%$ or mean accuracy $> 95\%$ are discarded, as they provide little information for differentiating between models.
    \item \textbf{Discrimination filtering:} We compute the point-biserial correlation $r_{pb}(i)$ between each item's response vector $\{Y_{i,\ell}\}_{\ell \in \mathcal{L}}$ and the models' total scores $T_\ell = \sum_{j \in \mathcal{I}} Y_{j,\ell}$ (see Appendix~\ref{app:preprocessing_details} for details). Items with $r_{pb}(i) < 0.1$ are removed as non-diagnostic.
\end{itemize}

This filtering process yields a refined response matrix that supports stable and reliable IRT calibration (see Table~\ref{tab:filtering_summary} for detailed results).

\subsection{Scalable IRT Calibration}\label{sec:calibration}

The calibration stage estimates item parameters \((a_i, b_i, c_i)\) and computes reference ability estimates \(\hat{\theta}_\ell^{\mathrm{whole}}\) for each LLM \(\ell\) for validation. To model the probability of a correct response, we adopt the three-parameter logistic (3PL) IRT model \citep{birnbaum1968some, lord1980}:

\begin{equation}
p_i(\theta_\ell) = c_i + \frac{1 - c_i}{1 + \exp(-a_i(\theta_\ell - b_i))}.
\end{equation}

Here, \(a_i\) is the discrimination parameter, controlling how sharply item \(i\) distinguishes between models with different latent abilities. 
\(b_i\) is the difficulty parameter, indicating the ability level at which the success probability reaches the midpoint between the lower asymptote \(c_i\) and perfect performance. 
\(c_i\) is the lower-asymptote parameter, accounting for the non-zero probability that lower-ability models answer an item correctly for unintended reasons. 
These parameters characterize the measurement quality of benchmark items and enable Fisher information-based item selection in our adaptive testing framework. 
Items with high discrimination and appropriate difficulty contribute more information about model ability, while weakly discriminative or saturated items are naturally down-weighted during evaluation.

\paragraph{Common-Person Calibration at Scale.} To estimate item characteristics efficiently using the 3PL model, we adopted a partition-based calibration procedure that leverages the unique structure of LLM benchmarking. Instead of fitting the full 3PL model to the entire item pool at once, which would be computationally prohibitive, we divided the items into $K$ non-overlapping subsets $\mathcal{I}_k$ (each with $|\mathcal{I}_k| \geq 100$ items), and calibrated each subset independently. This yields multiple provisional difficulty scales that must be aligned. Because all models answer all items, the model population serves as a natural set of common persons, allowing us to link the independently calibrated subsets onto a unified scale using mean–sigma transformations \citep{kolen2014}. This approach reduces computational complexity from $O(|\mathcal{I}|^3)$ to $O(K \cdot \max_k|\mathcal{I}_k|^3)$ while maintaining calibration accuracy due to the stability provided by having all models serve as linking anchors \citep{chalmers2012mirt}. A detailed description is provided in Appendix~\ref{common_person_calibration}.

\paragraph{Heterogeneity-Aware Ability Estimation.}
LLM populations exhibit extreme heterogeneity, ranging from near-random models ($\theta \approx -3$) to highly capable systems ($\theta \approx 3$). To obtain stable and unbiased estimates across this wide ability spectrum, we adopt the Weighted Likelihood Estimator (WLE) \citep{warm1989weighted}, which incorporates a bias-correction term $\frac{J(\theta)}{2I(\theta)}$, where $J(\theta) = \sum_i \frac{\partial I_i(\theta)}{\partial \theta}$. WLE provides finite, well-behaved estimates even at ability extremes and maintains desirable consistency properties \citep{baker2004item}. These characteristics are essential for establishing reliable evaluation baselines under the substantial heterogeneity present in modern LLM benchmarks.

\paragraph{Multi-Subset Model Fit Validation.}
Unlike prior IRT-based LLM benchmarks \citep{polo2024tinybenchmarks, metabench}, which do not report model-fit diagnostics, we perform explicit psychometric validation to ensure calibration quality. We compute the limited-information $M_2$ statistic with RMSEA \citep{maydeu2013goodness} for TinyBenchmarks, MetaBench, and ATLAS (Table~\ref{m2_compare}). TinyBenchmarks exhibits severe misfit across all datasets, consistent with its underidentified calibration setting—estimating up to 15 latent traits from only 395 models. MetaBench performs better (RMSEA 0.04–0.14) but still shows poor fit on TruthfulQA and marginal fit on ARC, indicating uneven generalization across domains. In contrast, ATLAS consistently achieves acceptable or good fit across all benchmarks, demonstrating stable and robust calibration. Our partition-based calibration further supports reliable validation by dividing the item bank into non-overlapping subsets (each $\geq$100 items) while using a shared model set as common persons. This linking strategy ensures that fit statistics reflect global calibration quality rather than subset-specific artifacts, which is essential for accurate Fisher-information–based adaptive testing.

\begin{table*}[ht]
\centering
\footnotesize

\begin{tabular}{l|cc|cc|cc|cc|cc}
\toprule
\multirow{2}{*}{\textbf{Method}}
& \multicolumn{2}{c|}{\textbf{Winogrande}} 
& \multicolumn{2}{c|}{\textbf{TruthfulQA}} 
& \multicolumn{2}{c|}{\textbf{HellaSwag}} 
& \multicolumn{2}{c|}{\textbf{GSM8K}} 
& \multicolumn{2}{c}{\textbf{ARC}} \\
 & RMSEA & Fit
 & RMSEA & Fit
 & RMSEA & Fit 
 & RMSEA & Fit 
 & RMSEA & Fit \\
\midrule
TinyBenchmarks  & 364.24  & Poor        & 371.49  & Poor        & 646.82  & Poor        & 506.60  & Poor        & 369.89  & Poor \\
MetaBench  & 0.0524  & Accept.  & 0.1389  & Poor        & 0.0498  & Good        & 0.0423  & Good        & 0.0811  & Marginal \\
ATLAS      & 0.0565  & Accept.  & 0.0690  & Accept.  & 0.0482  & Good        & 0.0438  & Good        & 0.0595  & Accept. \\
\bottomrule
\end{tabular}
\caption{
Model fit comparison across benchmarks using the limited-information statistic $M_2$ and its derived Avg. RMSEA values.
Lower Avg. RMSEA indicates better model fit. Model fit is interpreted according to standard psychometric thresholds: 
\textit{RMSEA $<$ 0.05 = Good fit; 0.05–0.08 = Acceptable fit (abbreviated as Accept.); 0.08–0.10 = Marginal fit; $>$ 0.10 = Poor fit. }
} 

\label{m2_compare}
\end{table*}
\subsection{Adaptive Testing with Information Selection}\label{sec:adaptive_testing}

Our proposed \CAT\ dynamically selects the most informative items for each model, dramatically reducing the number of items needed while maintaining accuracy. Algorithm~\ref{alg:run_cat} presents the complete procedure.
The algorithm includes several key design choices tailored to LLM evaluation:



\begin{algorithm}[h]
\caption{Adaptive Testing for Model $\ell$}
\label{alg:run_cat}
\begin{algorithmic}[1]
\STATE \textbf{Initialize:} $\hat{\theta}_0 \gets 0$, test record $R_\ell \gets \emptyset$, $t \gets 0$
\WHILE{$t < \mathrm{max\_items}$ and not converged}
    \STATE $t \gets t + 1$
    \IF{$t = 1$}
        \STATE Select item $i_t$ with $|b_{i_t} - \hat{\theta}_0|$ minimized
    \ELSE
        \STATE Compute Fisher information $I_i(\hat{\theta}_{t-1})$ for all unadministered items
        \STATE Select $i_t$ randomly from top-5 most informative items
    \ENDIF
    \STATE Administer item $i_t$ to model $\ell$, observe response $Y_{i_t,\ell}$
    \STATE Update record: $R_\ell \gets R_\ell \cup \{(i_t, Y_{i_t,\ell})\}$
    \STATE Update ability: $\hat{\theta}_t \gets \mathrm{EAP}(R_\ell)$
    \STATE Compute standard error: \\$\mathrm{SE}(\hat{\theta}_t) \gets 1/\sqrt{\sum_{j \in R_\ell} I_j(\hat{\theta}_t)}$
    \IF{$t \geq \mathrm{min\_items}$ and $\mathrm{SE}(\hat{\theta}_t) \leq  \tau$}
        \STATE \textbf{break} 
    \ENDIF
\ENDWHILE
\STATE \textbf{return} $\hat{\theta}_\ell \gets \hat{\theta}_t$, $\mathrm{SE}(\hat{\theta}_\ell)$, $R_\ell$
\end{algorithmic}
\end{algorithm}

\paragraph{Initialization and Bounds.} We initialize the ability estimate at $\hat{\theta}_0 = 0$. This value is a conventional neutral starting point in CAT because the latent trait scale is typically assumed to be centered at zero. Beginning at the scale midpoint helps stabilize early item selection by preventing the algorithm from drifting toward artificially high or low values before any response information is available. We enforce minimum (30) and maximum (500) item limits. The minimum ensures stable estimation for models at performance extremes, while the maximum constrains computational cost and yields approximately 90\% reduction in test length relative to full benchmarks.

\paragraph{Randomesque Item Selection.} Rather than deterministically selecting the single most informative item, we randomly sample from the top-5 candidates ranked by Fisher information:
\begin{equation}
I_i(\theta)
= \frac{\big(p_i'(\theta)\big)^2}
       {p_i(\theta)\,\big(1-p_i(\theta)\big)}
\end{equation}

This randomesque strategy \citep{kingsbury1989procedures} prevents over-reliance on specific item types while still keeping high information, which is important for models with specialized capabilities.

\paragraph{Sequential Ability Updates.} After each item administration, we update the ability estimate using Expected A Posteriori (EAP) estimation \citep{bock1982adaptive}:
\begin{equation}
\hat{\theta}_t = \mathbb{E}[\theta | R_\ell] = \int \theta \cdot p(\theta | R_\ell) \, d\theta .
\end{equation}
EAP provides numerically stable updates with sparse early responses and incorporates prior knowledge about ability distributions. In contrast, WLE tends to become unstable when response patterns are extreme, a situation common in the early stages of adaptive testing.

\paragraph{Precision-Based Stopping.}
In our implementation, testing stops once either the maximum item limit is reached or $\mathrm{SE}(\hat{\theta}_\ell)$ falls below a threshold $\tau$, after a minimum number of items has been administered.
 This precision-based rule ensures consistent measurement accuracy while minimizing test length. Although our experiments adopt this precision-based design, ATLAS can also operate under a fixed-length stopping rule by specifying a predetermined test length.

\paragraph{Output and Validation.} For each model $\ell$, the algorithm produces: (1) the administered item sequence and responses $R_\ell$, (2) the ability estimate trajectory $\{\hat{\theta}_t\}$ with associated standard errors, and (3) the final estimate $\hat{\theta}_\ell$. We validate these adaptive estimates against whole-bank references $\hat{\theta}_\ell^{\mathrm{whole}}$ to confirm that our dramatic reduction in items does not compromise measurement accuracy.
\begin{figure*}[ht]
    \centering
    \begin{subfigure}{0.325\textwidth}
        \centering
        \caption{ARC}
        \includegraphics[width=\linewidth]{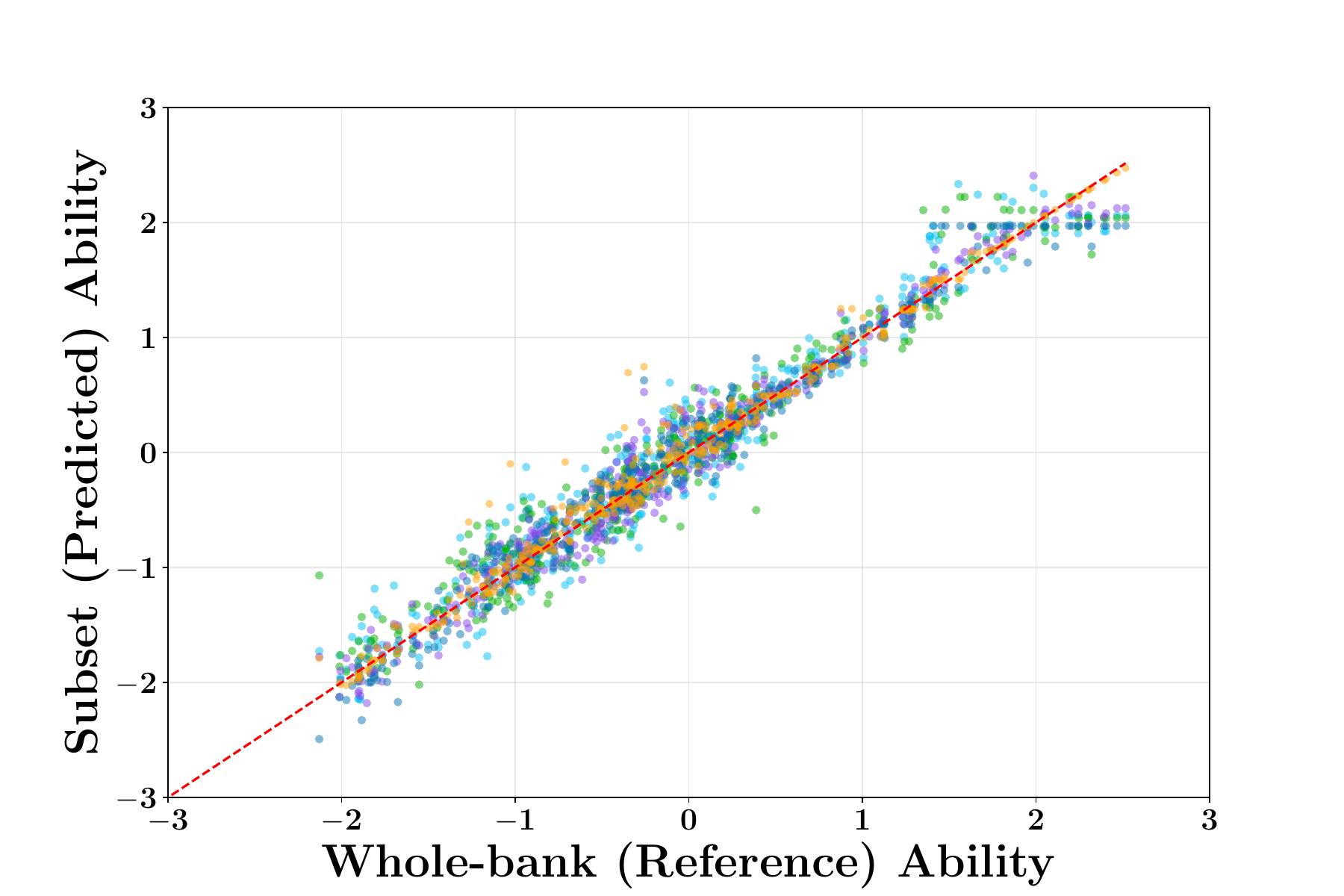}
    \end{subfigure}
    \begin{subfigure}{0.325\textwidth}
        \centering
        \caption{TruthfulQA}
        \includegraphics[width=\linewidth]{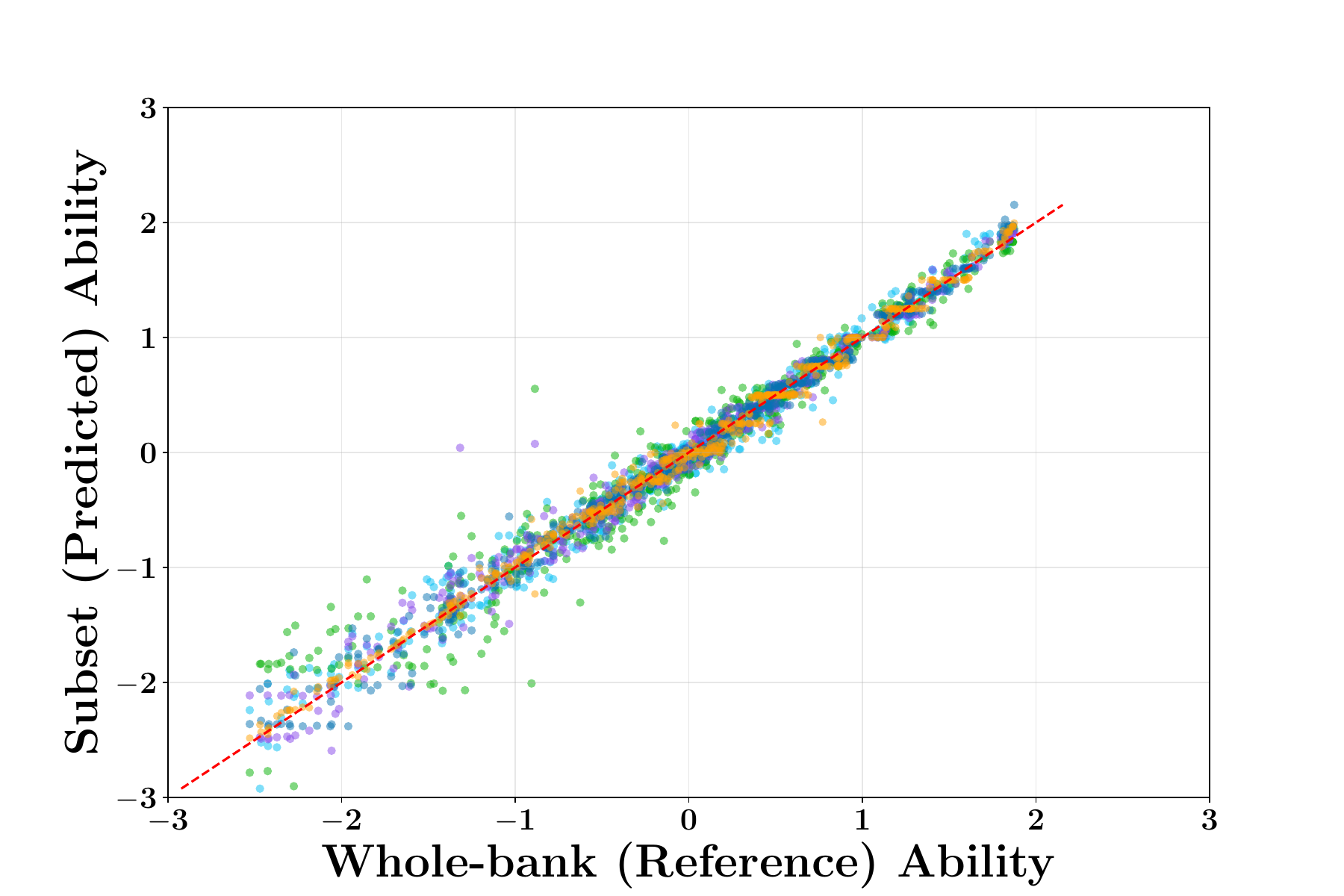}
        
    \end{subfigure}
    \begin{subfigure}{0.325\textwidth}
        \centering
        \caption{HellaSwag}
        \includegraphics[width=\linewidth]{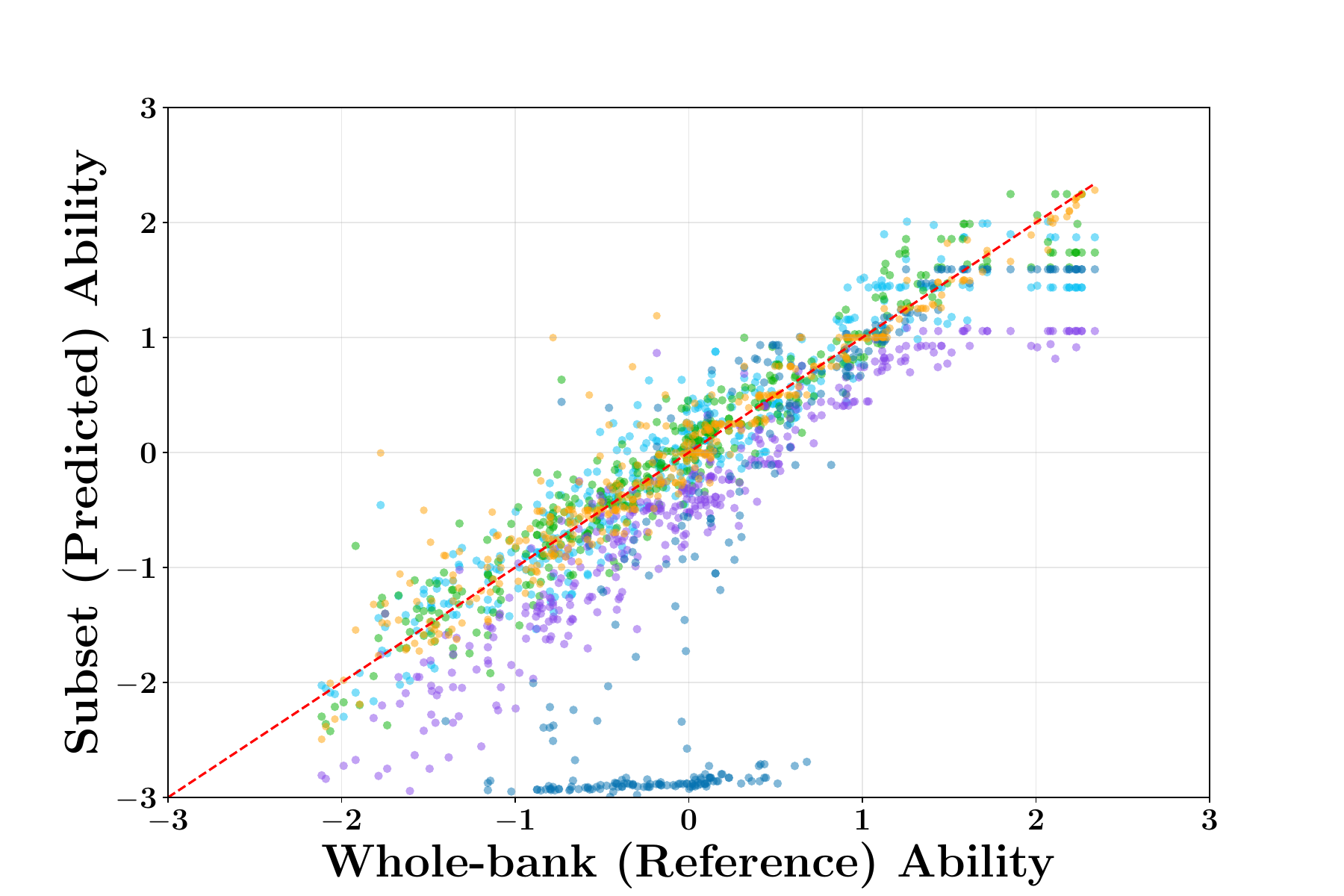}
    \end{subfigure}


    \begin{subfigure}{0.325\textwidth}
        \centering
        \caption{GSM8K}
        \includegraphics[width=\linewidth]{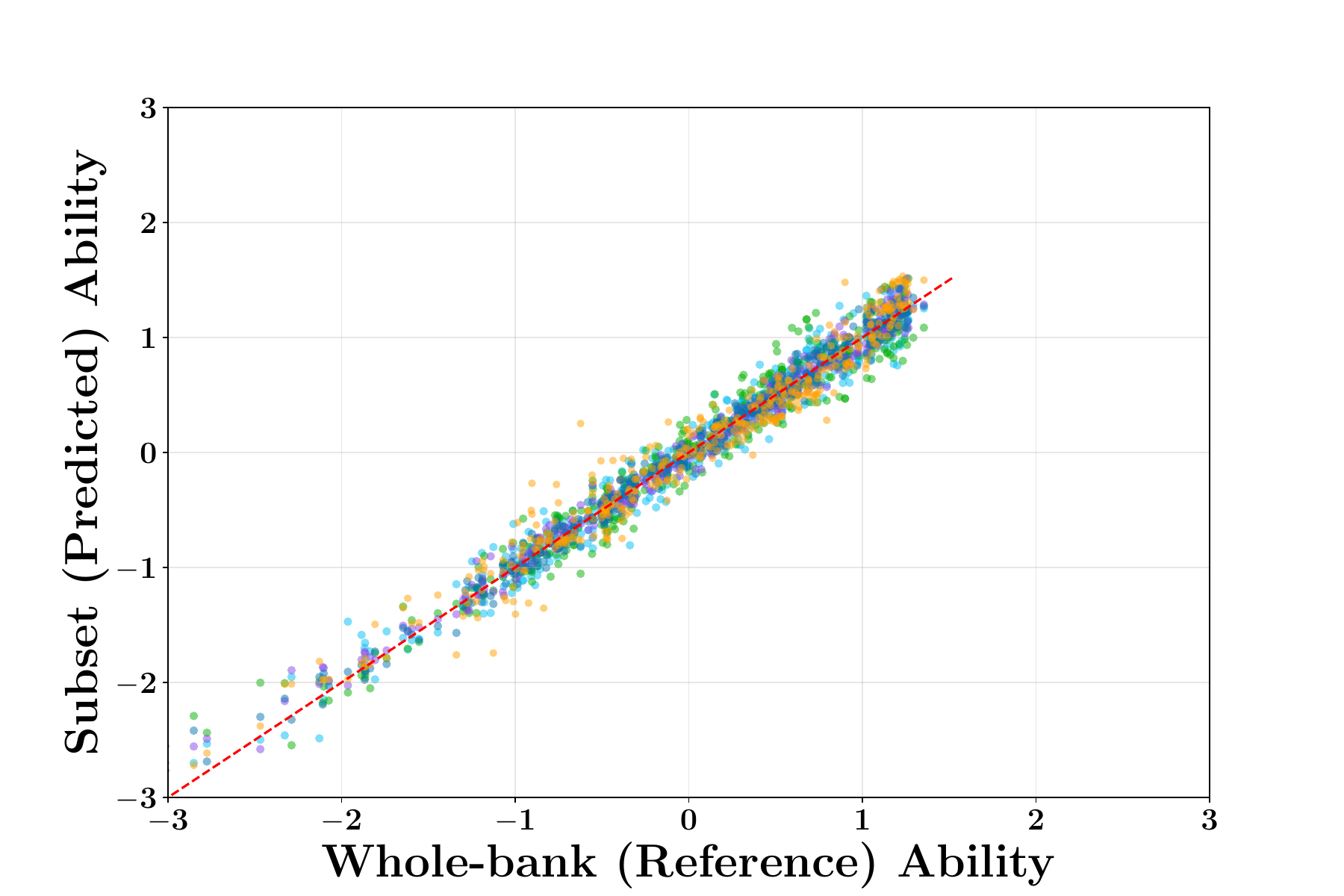}
    \end{subfigure}
    \begin{subfigure}{0.325\textwidth}
        \centering
        \caption{WinoGrande}
        \includegraphics[width=\linewidth]{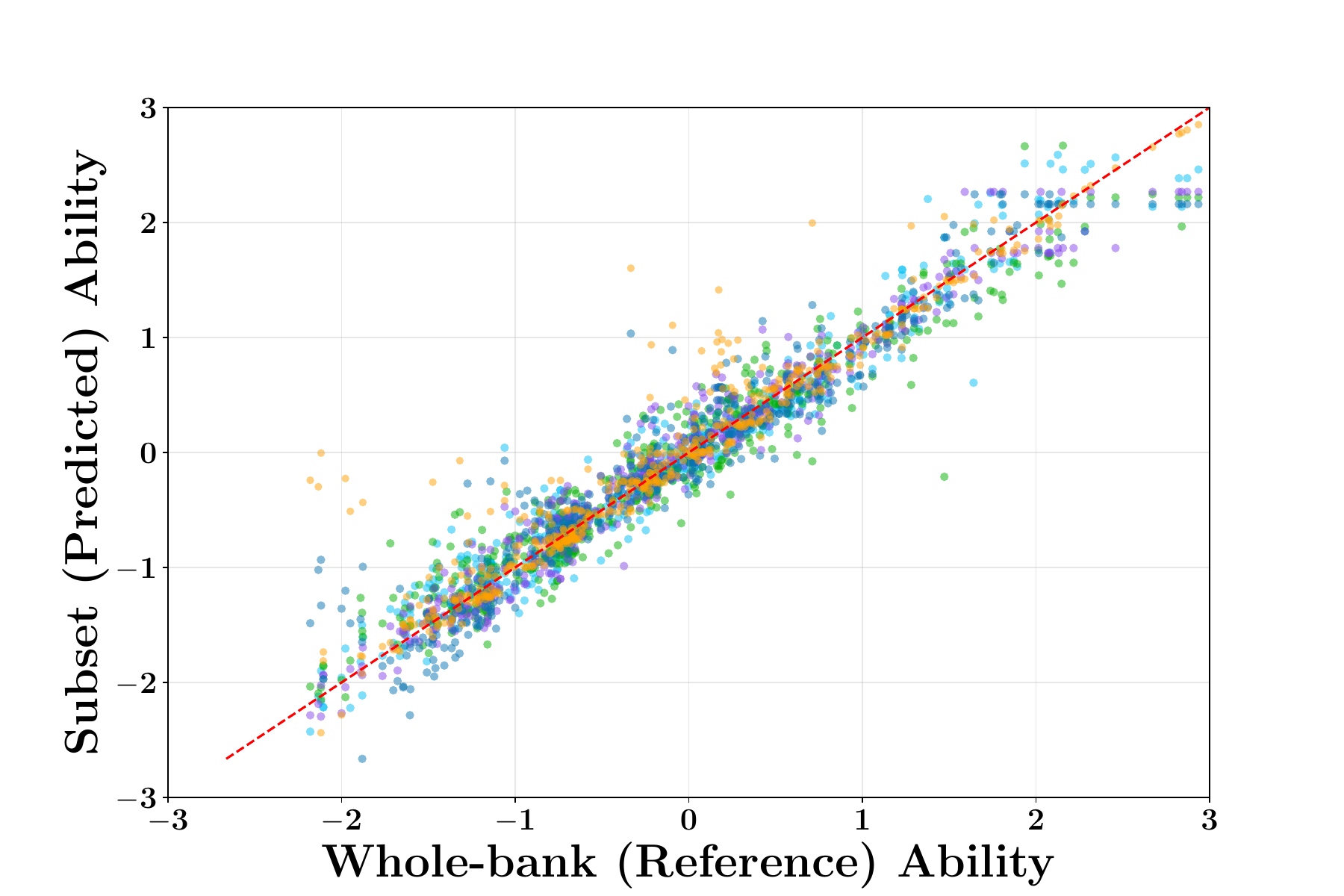}
    \end{subfigure}
    \begin{subfigure}{0.325\textwidth}
        \centering
        \includegraphics[width=\linewidth]{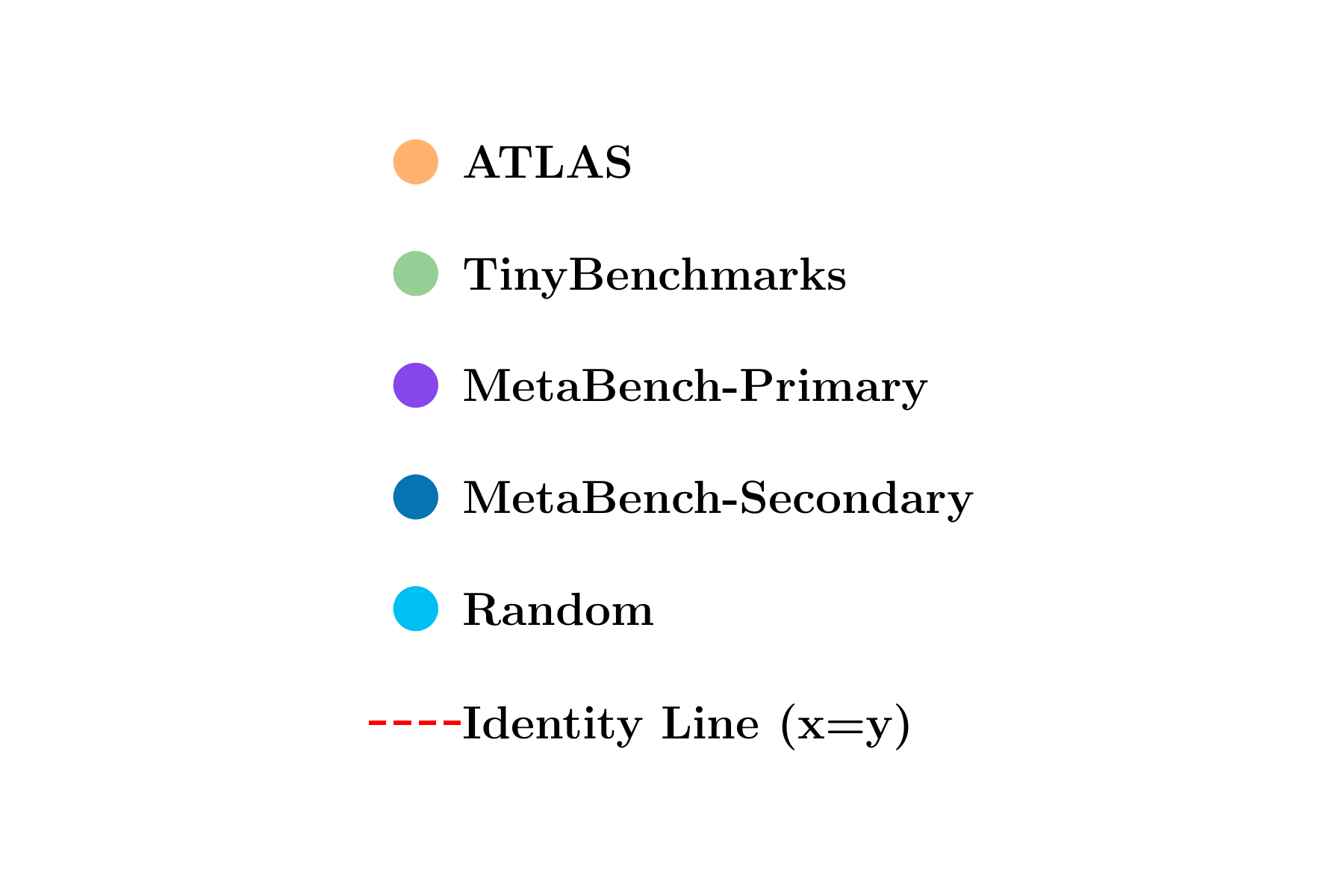}
    \end{subfigure}

    \caption{Comparison of subset (predicted) ability estimates against whole-bank (reference) abilities across five benchmarks (graphical illustration complementing Table~\ref{tab:main_result}). Points along the identity line indicate perfect agreement. \CAT\ maintains the closest alignment overall, particularly on TruthfulQA, ARC and HellaSwag and in the high-ability regime of WinoGrande, while static baselines such as TinyBenchmarks and Metabench show greater variance and systematic deviation.}
    \label{fig:theta_compare}
\end{figure*}

\begin{table*}[ht]
\centering
\small
\begin{tabular}{l|ccc|ccc|ccc}
\toprule
\multirow{2}{*}{\textbf{Method}} &
\multicolumn{3}{c|}{\textbf{WinoGrande}} &
\multicolumn{3}{c|}{\textbf{TruthfulQA}} &
\multicolumn{3}{c}{\textbf{HellaSwag}} \\
& MAE$\pm$SE $\downarrow$  & Items $\downarrow$  & IES $\downarrow$
& MAE$\pm$SE $\downarrow$  & Items $\downarrow$  & IES $\downarrow$
& MAE$\pm$SE $\downarrow$  & Items $\downarrow$  & IES $\downarrow$ \\
\midrule

Random$_{100}$
& 0.167$\pm$0.007 & 100 & 1.000
& 0.103$\pm$0.004 & 100 & 1.000
& 0.240$\pm$0.010 & 100 & 1.000 \\

TinyBenchmarks
& 0.204$\pm$0.008 & 100 & 1.221
& 0.145$\pm$0.007 & \dashuline{97} & 1.370
& 0.198$\pm$0.009 & 97 & 0.797 \\

MetaBench-P
& \textbf{0.152$\pm$0.007} & 133 & 1.216
& 0.084$\pm$0.004 & 154 & 1.262
& 0.514$\pm$0.016 & 93 & 1.990 \\

MetaBench-S
& 0.195$\pm$0.009 & 106 & 1.239
& \dashuline{0.072$\pm$0.003} & 136 & 0.945
& 1.570$\pm$0.055 & \dashuline{58} & 3.788 \\

ATLAS$_{0.1}$
& \underline{0.155$\pm$0.012} & \dashuline{70} & \dashuline{0.655}
& \textbf{0.064$\pm$0.002} & \underline{48} & \dashuline{0.300}
& \textbf{0.157$\pm$0.010} & \dashuline{41} & \dashuline{0.266} \\

ATLAS$_{0.2}$
& \dashuline{0.166$\pm$0.010} & \underline{37} & \underline{0.372}
& 0.073$\pm$0.003 & \textbf{30} & \underline{0.211}
& \underline{0.163$\pm$0.009} & \textbf{30} & \textbf{0.203} \\

ATLAS$_{0.3}$
& 0.179$\pm$0.011 & \textbf{32} & \textbf{0.342}
& \underline{0.071$\pm$0.003} & \textbf{30} & \textbf{0.206}
& \dashuline{0.165$\pm$0.010} & \textbf{30} & \underline{0.205} \\

\bottomrule
\end{tabular}
\caption{Comparison of whole-bank ability $\hat{\theta}_{\ell}^{\text{whole}}$ and subset-based
ability $\hat{\theta}_{\ell}$ across benchmarks. For each method, we report MAE$\pm$SE,
item count, and Information Efficiency Score (IES), where lower values are
better for all metrics. Bold indicates the best result, underlining the
second-best, and dashed underlining the third-best. Results for GSM8K and ARC are provided in Appendix~\ref{app:additional_tables}.}
\label{tab:main_result}
\end{table*}

\begin{table*}[ht]
\centering
\small

\begin{tabular}{l|ccc|ccc|ccc}
\toprule
\multirow{2}{*}{\textbf{Method}} &
\multicolumn{3}{c|}{\textbf{WinoGrande}} &
\multicolumn{3}{c|}{\textbf{TruthfulQA}} &
\multicolumn{3}{c}{\textbf{HellaSwag}} \\
& MAE$\pm$SE $\downarrow$  & Items $\downarrow$  & IES $\downarrow$
& MAE$\pm$SE $\downarrow$  & Items $\downarrow$  & IES $\downarrow$
& MAE$\pm$SE $\downarrow$  & Items $\downarrow$  & IES $\downarrow$ \\
\midrule

Random$_{100}$
& \underline{0.049$\pm$0.001} & 100 & 1.000
& \underline{0.021$\pm$0.001} & 100 & 1.000
& 0.024$\pm$0.001 & 100 & 1.000 \\

TinyBenchmarks
& \dashuline{0.050$\pm$0.001} & 100 & 1.010
& 0.025$\pm$0.001 & \dashuline{97} & 1.154
& \textbf{0.019$\pm$0.001} & 97 & 0.782 \\

MetaBench-P
& 0.054$\pm$0.001 & 133 & 1.446
& \textbf{0.017$\pm$0.001} & 154 & 1.266
& 0.050$\pm$0.001 & 93 & 1.943 \\

MetaBench-S
& 0.051$\pm$0.001 & 106 & 1.103
& \underline{0.021$\pm$0.001} & 136 & 1.394
& 0.115$\pm$0.004 & \dashuline{58} & 2.793 \\

ATLAS$_{0.1}$
& \textbf{0.048$\pm$0.001} & \dashuline{70} & \dashuline{0.678}
& \dashuline{0.023$\pm$0.001} & \underline{48} & \dashuline{0.532}
& \underline{0.020$\pm$0.001} & \underline{41} & \dashuline{0.348} \\

ATLAS$_{0.2}$
& 0.051$\pm$0.002 & \underline{37} & \underline{0.383}
& 0.024$\pm$0.001 & \textbf{30} & \underline{0.338}
& \dashuline{0.021$\pm$0.001} & \textbf{30} & \textbf{0.258} \\

ATLAS$_{0.3}$
& \dashuline{0.050$\pm$0.001} & \textbf{32} & \textbf{0.324}
& \dashuline{0.023$\pm$0.001} & \textbf{30} & \textbf{0.331}
& \dashuline{0.021$\pm$0.001} & \textbf{30} & \underline{0.261} \\

\bottomrule
\end{tabular}
\caption{Comparison of raw whole-bank accuracy and p-IRT reconstructed accuracy across benchmarks.
For each method, we report MAE$\pm$SE, number of administered items, and the
Information Efficiency Score (IES), where lower values are
better for all metrics. Bold denotes the best value,
underlining the second-best, and dashed underlining the third-best. Results for GSM8K and ARC are provided in Appendix~\ref{app:additional_tables}.}
\label{tab:acc_results}
\end{table*}

\section{Experiments}\label{sec:results}

We evaluate the proposed \CAT\ framework across five benchmarks, comparing its efficiency and accuracy against static baselines such as random sampling, TinyBenchmarks \citep{polo2024tinybenchmarks}, MetaBench-Primary and MetaBench-Secondary \citep{metabench}. We report performance-based and efficiency-based metrics, with full metric definitions provided in Appendix~\ref{app:metrics}.

\subsection{Experimental Setup and Metrics}

We evaluate \CAT\ across five diverse benchmarks covering different cognitive domains: WinoGrande (commonsense reasoning), TruthfulQA (factual consistency), HellaSwag (procedural inference), GSM8K (mathematical reasoning), and ARC (scientific question answering). All experiments use calibrated item banks from Section~\ref{sec:calibration}. 

We compare against four static baselines that do not adapt to individual models: (1) \textbf{Random sampling} of 100 items uniformly from the full bank, (2) \textbf{TinyBenchmarks} \citep{polo2024tinybenchmarks} using predetermined subsets selected via clustering without Fisher information optimization, (3) \textbf{MetaBench-Primary} and (4) \textbf{MetaBench-Secondary} \citep{metabench} using curated splits that require computationally expensive iterations to identify stable subsets. Unlike these static approaches, \CAT\ uses three precision thresholds ($\mathrm{SE}(\hat{\theta}) \leq 0.1, 0.2, 0.3$) and an item bound of 30--500, terminating adaptively when the required precision is achieved or the maximum test length is reached. Additional experimental configurations are provided in Appendix~\ref{app:experimental_details}.

We evaluate each method primarily in \textbf{ability space}, comparing ATLAS-estimated abilities $\hat{\theta}_\ell$ with full-bank abilities $\hat{\theta}^{\text{whole}}_\ell$ (See Table~\ref{tab:main_result}). We report four metrics: 
(1) \textbf{Mean Absolute Error (MAE)} to measure estimation accuracy; 
(2) \textbf{Standard Error (SE)} of the absolute errors across models to quantify stability; 
(3) \textbf{Average Test Length}, the number of items administered per model; and 
(4) \textbf{Information Efficiency Score (IES)}, which jointly reflects accuracy and item usage relative to a 100-item random baseline (values $< 1$ indicate higher efficiency). More details of these metrics are provided in Appendix~\ref{app:metrics}. We also provide \textbf{accuracy-space evaluations}, comparing reconstructed accuracies to full-bank raw accuracies (See Table~\ref{tab:acc_results}). Additional evaluation metrics and their results, including Item Exposure Rate, Test Overlap Rate, and Selection Time are provided in Appendix~\ref{app:table6_analysis}.

\begin{figure}[ht]
    \centering
    \includegraphics[width=\linewidth]{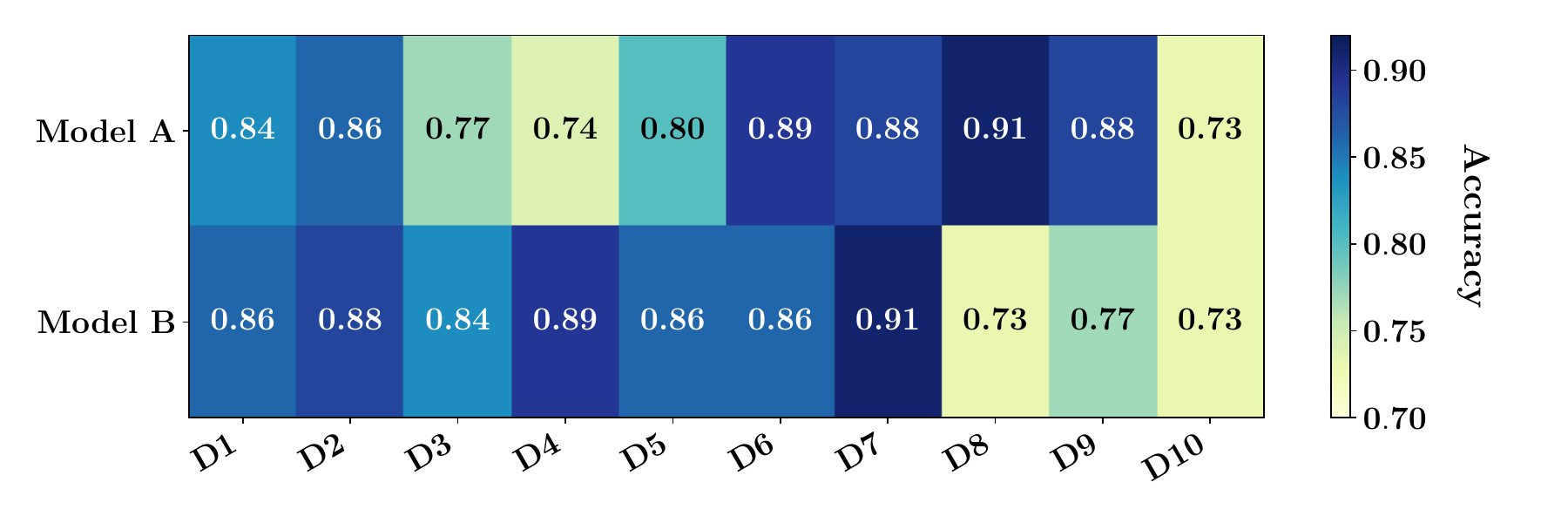}
    \caption{Two models with identical accuracy (0.833) on WinoGrande receive different ability estimates ($\hat{\theta}_A = 1.2$ vs $\hat{\theta}_B = 0.6$). Difficulty levels are ordered from easiest (D1) to hardest (D10). Model A performs better on harder items (darker cells on right), while Model B primarily succeeds on easier ones (darker cells on left), highlighting how IRT distinguishes latent ability differences that accuracy fails to capture.}
    \label{fig:diff_same_acc}
\end{figure}

\begin{figure*}[ht]
    \centering
    \begin{subfigure}{0.48\textwidth}
        \centering
                \caption{GSM8K}
        \includegraphics[width=\linewidth]{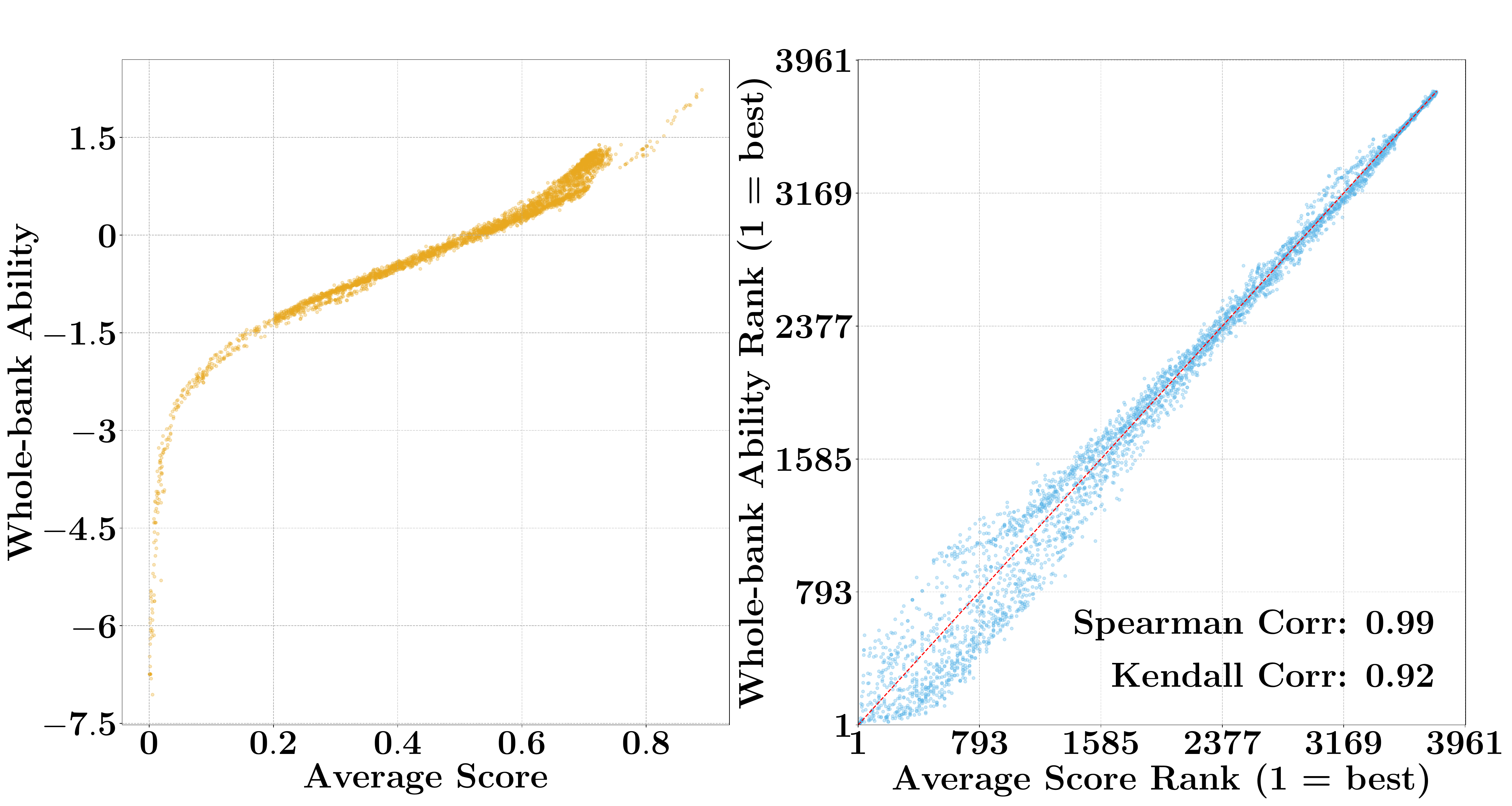}

    \end{subfigure}
    \begin{subfigure}{0.48\textwidth}
        \centering
        \caption{HellaSwag}
        \includegraphics[width=\linewidth]{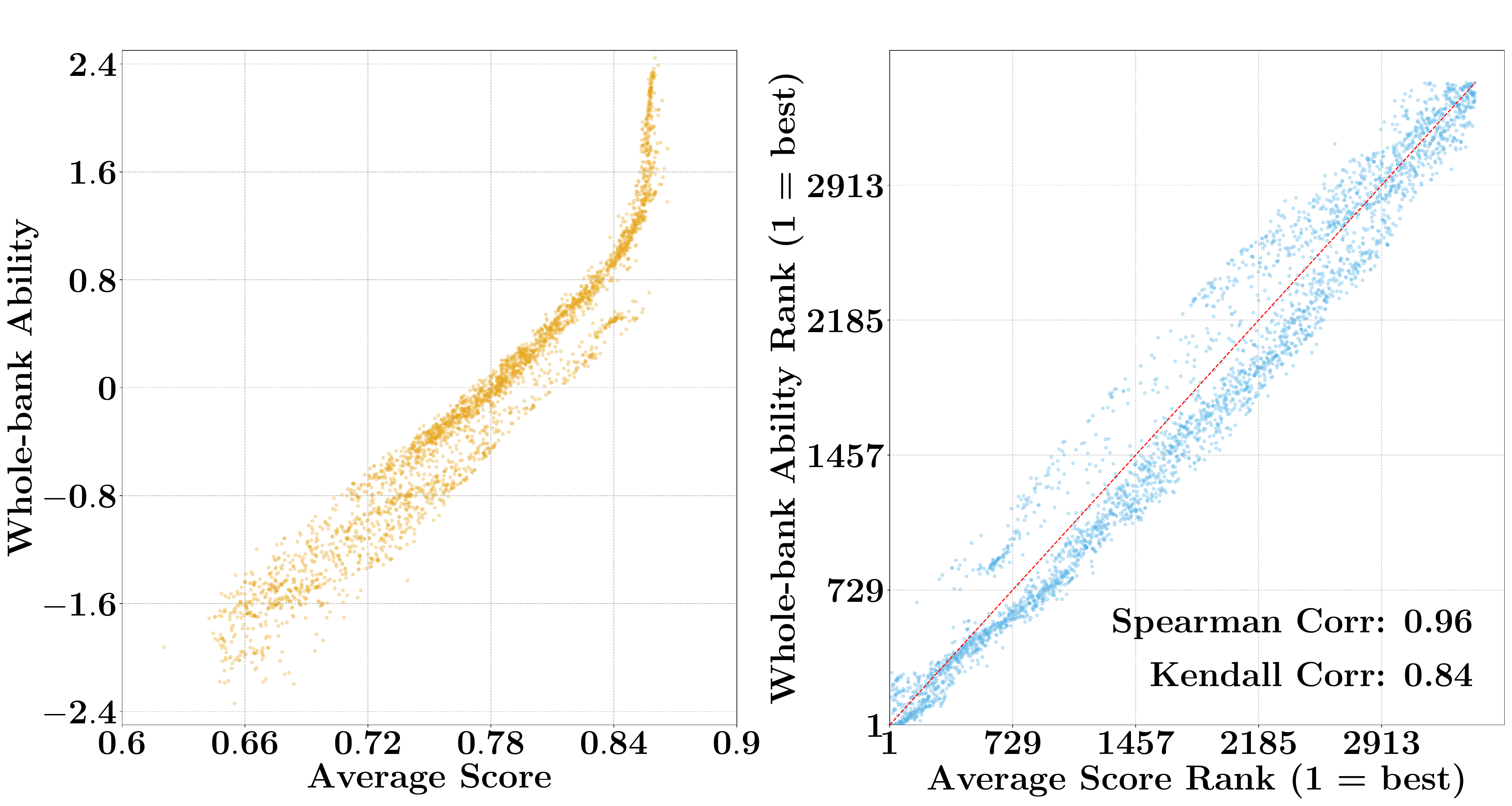}

    \end{subfigure}

\caption{Comparison of IRT ability estimates $\hat{\theta}_\ell^{\text{whole}}$ with raw accuracy. Left: Ability vs accuracy reveals strong correlation but critical differences at performance extremes where accuracy collapses. Right: Rank comparison shows systematic reordering, with 23\% (GSM8K) and 31\% (HellaSwag) of models shifting $>$ 10 positions. IRT separates models with identical accuracies by accounting for which items they solve correctly.}

\label{fig:score_vs_theta}
\end{figure*}

\subsection{Performance and Reliability Analysis}
Table~\ref{tab:main_result} presents a comparison of whole-bank ability estimates and subset-based estimates across benchmarks (results for GSM8K and ARC are provided in Appendix~\ref{app:additional_tables}). ATLAS consistently delivers the strongest accuracy–efficiency tradeoff among all methods. Across every benchmark, an ATLAS variant achieves the lowest Information Efficiency Score (IES), indicating that it achieves the best accuracy-efficiency trade-off. For example, ATLAS attains the best MAE on TruthfulQA (0.064 with 48 items) and HellaSwag (0.157 with 41 items), and matches the performance of MetaBench-Primary on WinoGrande while requiring nearly half as many items (70 vs. 133). Even in more challenging settings such as GSM8K and ARC (Appendix~\ref{app:additional_tables}), ATLAS maintains low MAE with item counts as small as 30–36, outperforming all static baselines in information efficiency. In contrast, static subsets such as TinyBenchmarks and MetaBench exhibit inconsistent performance. They are strong on some datasets but poor on others, with substantially higher IES values. Overall, these results demonstrate that adaptive item selection enables ATLAS to deliver high-fidelity ability estimates while dramatically reducing evaluation cost, achieving reliable performance that fixed subsets fail to match.

Figure \ref{fig:theta_compare} plots subset-based ability estimates against whole-bank references across all five benchmarks. Across datasets, ATLAS shows the closest alignment to this identity line, with tightly clustered points and minimal systematic drift, reflecting stable and high-fidelity ability estimation. In contrast, static baselines exhibit benchmark-dependent failures. Random sampling yields the widest scatter on every benchmark. TinyBenchmarks consistently deviate from the identity line in the extreme high or low-ability regime, especially on ARC and TruthfulQA. MetaBench-Primary and MetaBench-Secondary performs reasonably on GSM8K but breaks down substantially on HellaSwag, where both its primary and secondary subsets produce large downward deviations, indicating poor item coverage. 

As shown in Table~\ref{tab:results_test_overlap}, \CAT\ produces adaptive tests with low redundancy and high efficiency. Test overlap rates remain between 11–23\%, and item exposure stays below 12\% across all benchmarks, reflecting broad utilization of the item bank. End-to-end runtimes range from 9.4 to 75.5 seconds per model and scale consistently with bank size (shortest for TruthfulQA and longest for HellaSwag), confirming that \CAT\ is computationally practical.



\paragraph{Accuracy Reconstruction.}
To assess whether the estimated ability $\theta$ is consistent with traditional accuracy-based evaluation, we reconstruct each model’s expected percent correct from its estimated ability and observed responses on the reduced subset using the \textit{performance-IRT} (p-IRT) estimator \citep{polo2024tinybenchmarks} (detailed in Appendix~\ref{app:p-irt}). P-IRT builds on the Test Characteristic Curve \citep{lord2008statistical, hambleton1991fundamentals} which maps a model’s ability $\hat{\theta}$ to its expected probability of correctly answering benchmark items under the calibrated 3PL model, and refines this mapping by incorporating observed responses.

As shown in Table~\ref{tab:acc_results}, across all benchmarks, the reconstructed percent correct closely matches the raw full-bank percent correct, with MAE consistently below 5\%, demonstrating that $\theta$ faithfully preserves global performance as measured by accuracy. This confirms that the latent ability estimates remain well aligned with conventional evaluation metrics.
 Moreover, ATLAS achieves this high reconstruction fidelity with substantially fewer administered items than competing methods. For example, on WinoGrande, ATLAS$_{0.1}$ attains the lowest MAE (0.048) using only 70 items, compared to 133 items for MetaBench-Primary. 

\paragraph{Sensitivity to the IRT parameterization.}
We use the 3PL model as the default IRT parameterization and compare it against a 2PL variant that omits the lower-asymptote parameter  \(c_i\) . As shown in Table~\ref{tab:2pl3pl_combined}, the benefit of 3PL is benchmark-dependent rather than universal.
On TruthfulQA, 3PL improves both ability estimation and accuracy reconstruction while using similar or fewer items. On WinoGrande, GSM8K, ARC, and HellaSwag, results are more mixed: 2PL sometimes achieves slightly lower ability MAE, but often requires substantially longer tests, whereas 3PL remains competitive and usually with better accuracy reconstruction. Full results are provided in Appendix~\ref{app:2pl3pl}.

\subsection{Items Are Not Equally Informative}
Figure~\ref{fig:diff_same_acc} illustrates a core principle of Item Response Theory: items contribute unequally to ability estimation. Although Models A and B achieve the same raw accuracy (0.833), their response patterns differ systematically across difficulty levels, leading to substantially different latent ability estimates ($\hat{\theta}_A = 1.2$ vs.\ $\hat{\theta}_B = 0.6$). Model A performs disproportionately well on harder items (D6–D9), which are more discriminative and thus carry greater informational value for estimating high ability, whereas Model B’s correct responses are concentrated on easier items (D1–D4), which provide limited evidence of advanced competence. As a result, \textbf{identical accuracy masks meaningful differences in underlying capability}: two models can answer the same number of questions correctly, yet differ substantially in latent ability because correct responses to difficult items provide strong evidence of high ability, while correct responses to easy items are expected and contribute little additional information. This example highlights why accuracy alone is an insufficient evaluation metric and how IRT leverages item difficulty to recover latent ability differences that raw accuracy fail to capture. Similar patterns of identical accuracy leading to different ability estimates are observed across ARC, HellaSwag, and other benchmarks, with additional heatmap visualizations provided in Appendix~\ref{app:additional_results}.

\subsection{Discriminating Models at Performance Extremes}
Figure~\ref{fig:score_vs_theta} illustrates IRT’s key advantage: separating models with similar accuracy via latent ability estimates. Although accuracy and ability are highly correlated (0.99 on GSM8K; 0.96 on HellaSwag), systematic differences appear at performance extremes. In low-performing regimes, accuracy collapses into narrow bands (0.10–0.15), while IRT spans $\theta \approx -3$ to $-1$, distinguishing models that solve only easy items from those that handle harder ones. At the high end, ceiling effects compress accuracy, whereas IRT preserves discrimination across $\theta \approx 1.5$ to $2.5$.

The right panels reveal substantial rank reordering: 23–31\% of models shift by more than 10 positions when ranked by IRT rather than accuracy. Models that succeed on difficult items are ranked higher, while those relying on easy items are downweighted, yielding more reliable comparisons in saturated regimes. Similar trends hold across TruthfulQA, WinoGrande, and ARC (See Appendix~\ref{app:additional_results}).

\begin{table}[t]
\centering
\small
\setlength{\tabcolsep}{5pt}
\begin{tabular}{lccc}
\toprule
\textbf{Setting} & \textbf{Raw Acc.} & \textbf{Ability} & $\boldsymbol{\Delta}$ \\
\midrule

\multicolumn{4}{l}{\textit{Split-half ranking stability (Spearman correlation $\uparrow$)}} \\
WinoGrande & 0.943 & \textbf{0.981} & +0.038 \\
TruthfulQA & 0.981 & \textbf{0.992} & +0.011 \\
GSM8K & 0.985 & \textbf{0.993} & +0.008 \\
ARC & 0.968 & \textbf{0.977} & +0.008 \\
HellaSwag & 0.994 & \textbf{0.996} & +0.002 \\

\midrule

\multicolumn{4}{l}{\textit{Related-benchmark ranking consistency (Spearman correlation $\uparrow$)}} \\

\begin{tabular}[c]{@{}l@{}}
GSM8K $\leftrightarrow$ \\
MMLU Elementary Math
\end{tabular}
& 0.504 & \textbf{0.738} & +0.234 \\

\begin{tabular}[c]{@{}l@{}}
MMLU College Chem. $\leftrightarrow$ \\
MMLU High School Chem.
\end{tabular}
& 0.439 & \textbf{0.904} & +0.465 \\

\bottomrule
\end{tabular}
\caption{
Comparison of raw accuracy-based and ability-based rankings.
Split-half stability results are averaged over 10 random partitions.
Higher Spearman correlation is better. Ability rankings are more stable across benchmark forms
and more consistent across related benchmarks targeting similar abilities.
}
\label{tab:theta_stability_transfer}
\end{table}



\subsection{Ability-Based Rankings Are More Stable}

The ranking differences between raw accuracy and ability in Figure~\ref{fig:score_vs_theta} raise an important question: do ability-based rankings capture meaningful latent ability differences, or simply introduce additional estimation variance? To evaluate ranking stability, we compare model rankings across independently sampled subsets of benchmark items.

For each benchmark, we perform a repeated split-half stability analysis by randomly partitioning evaluation items into two disjoint subsets, A and B. Using each subset independently, we compute model rankings based on both raw accuracy and IRT ability estimates (\(\hat{\theta}_\ell^{\text{whole}}\)), and measure agreement using Spearman correlation. This procedure is repeated over 10 random partitions, and correlations are averaged across runs. 
As shown in Table~\ref{tab:theta_stability_transfer}, ability-based rankings consistently achieve higher split-half agreement across all five benchmarks. For example, ability-based ranking improves rank consistency from 0.943 to 0.981 on WinoGrande and from 0.985 to 0.993 on GSM8K, suggesting that the observed reorderings reflect a more stable latent ability signal rather than noisy reshuffling.

We further evaluate whether rankings transfer across different but related benchmarks, providing external evidence beyond same-benchmark stability. Ability-based rankings show substantially stronger alignment than raw-accuracy rankings between GSM8K and MMLU Elementary Mathematics (0.738 vs.\ 0.504), as well as between MMLU College Chemistry and MMLU High School Chemistry (0.904 vs.\ 0.439). These results suggest that IRT ability estimates better preserves shared domain-relevant ability structure across related benchmarks, consistent with prior findings on the robustness of IRT-based evaluation~\citep{rodriguez2021evaluation}.






\begin{table}[t]
\centering
\small
\setlength{\tabcolsep}{4pt}

\begin{tabular}{llcc}
\toprule
\multirow{2}{*}{\textbf{Method}} 
& \multirow{2}{*}{\textbf{Setting}}
& \textbf{Ability} 
& \textbf{Accuracy} \\
& & MAE$\pm$SE $\downarrow$ 
& MAE$\pm$SE $\downarrow$ \\
\midrule

\multirow{3}{*}{ATLAS$_{0.1}$}
& Random Split
& \textbf{0.084$\pm$0.006}
& \textbf{0.032$\pm$0.002} \\

& Family Split
& 0.091$\pm$0.005
& 0.044$\pm$0.002 \\

& Temporal Split
& 0.126$\pm$0.012
& 0.044$\pm$0.002 \\

\midrule

\multirow{3}{*}{ATLAS$_{0.2}$}
& Random Split
& 0.120$\pm$0.008
& \textbf{0.034$\pm$0.002} \\

& Family Split
& \textbf{0.107$\pm$0.006}
& 0.045$\pm$0.002 \\

& Temporal Split
& 0.157$\pm$0.013
& 0.044$\pm$0.002 \\

\midrule

\multirow{3}{*}{ATLAS$_{0.3}$}
& Random Split
& 0.117$\pm$0.007
& \textbf{0.034$\pm$0.002} \\

& Family Split
& \textbf{0.109$\pm$0.006}
& 0.044$\pm$0.002 \\

& Temporal Split
& 0.162$\pm$0.011
& 0.045$\pm$0.002 \\

\bottomrule
\end{tabular}

\caption{
Robustness of ATLAS under architecture-family and temporal
distribution shift on ARC. Lower values are better.
}

\label{tab:ood_generalization}
\end{table}

\subsection{Generalization Beyond the Calibration Distribution}

We evaluate whether ATLAS generalizes to models outside the calibration distribution under two settings on ARC: (1) strict architecture-family holdout and (2) temporal generalization across model release dates.

\paragraph{Architecture-family holdout.} 
We evaluate a strict family-holdout setting where all Mixtral-family models (321 models), a structurally distinct mixture-of-experts (MoE) architecture family, are withheld during calibration and evaluated only at test time. Item parameters are calibrated using the remaining 3,296 models. As shown in Table~\ref{tab:ood_generalization}, ATLAS transfers reasonably well to the held-out family. Raw-accuracy reconstruction MAE increases only modestly from 0.032--0.034 to 0.044--0.045, while ability estimation remains stable and even slightly improves at larger stopping thresholds.

\paragraph{Temporal generalization.}
We further evaluate temporal robustness by calibrating ARC on 2,998 models released before 2024-05-01 and evaluating on 322 later-release models (844 models were excluded due to missing release dates). As shown in Table~\ref{tab:ood_generalization}, ability estimation MAE increases from 0.084--0.117 to 0.126--0.162, while raw-accuracy reconstruction changes more modestly from 0.032--0.034 to 0.044--0.045. Overall, calibration remains reasonably transferable to newer model distributions.

\section{Discussion}
Our results underscore several psychometric principles that should guide the design of future LLM benchmarks. First, \textbf{item quality} is critical for reliable evaluation. Prior work has shown that mislabeled or ambiguous items are common in existing benchmarks \citep{vendrow2025large, gema2025we}. 
Although IRT naturally downweights such items through probabilistic modeling, static benchmarks lack mechanisms to prevent them from being sampled. Fluid Benchmarking \citep{hofmann2025fluid} highlights this issue, showing that adaptive, information-based selection dramatically reduces the exposure of low-quality items. These findings collectively reinforce the need for psychometric validation in future benchmark design, including procedures such as discrimination screening and content-alignment checks.



Second, \textbf{model fit} is essential for trustworthy item calibration but is rarely reported in benchmark-reduction pipelines that apply IRT. Poor model fit yields unstable difficulty and discrimination estimates, undermining downstream conclusions. Our analysis shows that several widely used static subsets exhibit substantial misfit (Table~\ref{m2_compare}), suggesting that model-fit diagnostics should be a standard requirement whenever IRT is used for item calibration or reduction.

Third, when large item banks are partitioned for computational efficiency, \textbf{linking} becomes necessary to prevent scale drift. Common-person linking places items on a shared latent scale, preserving interpretability and comparability. Such principled linking strategies are crucial for scalable benchmarking and for maintaining consistency across bank updates or domains.

Finally, reduced benchmarks require explicit \textbf{content balancing} to maintain validity. Without constraints on domain coverage, reduced item sets may overemphasize narrow skills. Our adaptive framework naturally supports joint optimization of measurement precision and content balance \citep{cheng2009maximum}, whereas achieving similar guarantees in static subsets  requires extensive manual tuning.




\section{Conclusion}
We presented ATLAS, an adaptive testing framework that reframes LLM
evaluation by moving beyond 
static, accuracy-based benchmarking toward
dynamic ability estimation. Through psychometric
modeling and information-guided item selection, ATLAS achieves up to 90\%
item reduction, avoids accuracy ceiling effects, and reveals ability differences that
static benchmarks overlook. Our analysis further
highlights the importance of rigorous model-fit validation, item-quality assessment,
and principled linking procedures for building reliable benchmarks, offering a more efficient, interpretable, and robust foundation to assess LLMs.


\section*{Impact Statement}

This work aims to advance the methodology of large language model evaluation by
introducing a psychometrically grounded, adaptive testing framework. By improving
the reliability, efficiency, and interpretability of model assessment, our approach
may support more informed benchmarking and comparison of LLMs. We do not foresee
direct negative societal or ethical impacts arising from this work, as it focuses on
evaluation rather than deployment or decision-making. Nonetheless, as with all
benchmarking tools, care should be taken to ensure that evaluations are used
responsibly and do not incentivize narrow optimization at the expense of broader
model robustness.

\nocite{langley00}

\bibliography{iclr2026_conference}
\bibliographystyle{icml2026}

\newpage
\appendix
\onecolumn
\appendix
\section{Detailed Analysis of Average Score Limitations}\label{app:avg_score_limits}

Average score (percent correct) remains the most widely reported metric for
evaluating LLMs. While it provides a convenient ordinal indicator for fixed
forms, it is a shaky measure of underlying ability.

First, average scores are \emph{form-dependent}: changing the mix or
difficulty of items alters percent correct, even if the model's true ability
is unchanged. Second, the metric has a \emph{nonlinear scale}: improvements at
the extremes (e.g., $98\%\!\to100\%$) do not reflect the same underlying gain
as improvements in the middle (e.g., $50\%\!\to52\%$). Third, it assumes
\emph{equal informativeness} across items, allowing easy or guessable items to
influence the mean as much as highly discriminative ones. Fourth, it is
subject to \emph{coverage bias}: the observed score reflects the content
blueprint of the test rather than ability across domains. Fifth, average
scores offer \emph{no measure of uncertainty}, making it unclear whether
differences are statistically meaningful. Finally, they are highly
\emph{sensitive to contamination}: memorized items from pretraining can
artificially inflate percent correct without reflecting genuine reasoning or
generalization.

In contrast, IRT-based ability estimates ($\theta$) provide form-invariant,
uncertainty-aware measures that adjust for item difficulty and discrimination.
Reporting $\theta \pm \text{SE}(\theta)$ offers a psychometrically principled
alternative. For communication purposes, reconstructed percent scores may be
shown alongside, but $\theta$ should serve as the primary indicator of model
capability.

\section{Comparison of IRT-Based Benchmark Methods}
\label{app:method_comparison}
A detailed comparison of IRT-based benchmark methods is reported in Table~\ref{tab:method_comparison}.

\begin{table*}[htbp]
\centering
\renewcommand{\arraystretch}{1.2}

\begin{tabular}{llll}
\toprule
\textbf{Factor} & \textbf{TinyBenchmarks (Static)} & \textbf{MetaBench (Static)} & \textbf{\CAT} \\
\midrule
IRT Calibration & Required & Required & Required \\
\midrule

Adaptivity & None (same items) & None (same items) & High (items vary by ability) \\
\midrule
Test Length & Fixed & Fixed & Variable, stopping rules \\
\midrule

Exposure control & High (same items reused) & High (same items reused) & Low (rotating pool) \\
\midrule
Pool Sensitivity & Subset dependent & Subset dependent & Robust to large pools \\
\midrule

Fairness & Biased if mis-targeted & Biased if mis-targeted & Balanced across abilities \\
\midrule
Score Precision & Low at extremes & Low at extremes & High, SEs available \\
\midrule

Model Fit & Rarely checked & Rarely checked & Possible fit checks \\
\midrule

Saturation Risk & High  & High & Low  \\
\bottomrule
\end{tabular}
\caption{Comparison of IRT-Based Benchmark Methods for LLMs}
\label{tab:method_comparison}
\end{table*}

\section{Data Preprocessing Details}
\label{app:preprocessing_details}

\subsection{Point-Biserial Correlation Formula}

The point-biserial correlation \citep{allen2001introduction} for item $i$ is defined as:
\[
r_{pb}(i) \;=\; \frac{\bar{T}_{\,\ell \mid Y_{i\ell}=1} - \bar{T}_{\,\ell \mid Y_{i\ell}=0}}{s_T} \cdot \sqrt{p_i q_i},
\]
where $\bar{T}_{\,\ell \mid Y_{i\ell}=1}$ and $\bar{T}_{\,\ell \mid Y_{i\ell}=0}$
are the mean total scores of models that answered item $i$ correctly and
incorrectly, respectively; $s_T$ is the standard deviation of total scores
$\{T_\ell\}$; $p_i = \tfrac{1}{|\mathcal{L}|} \sum_{\ell \in \mathcal{L}} Y_{i\ell}$
is the proportion of models that answered item $i$ correctly; $q_i = 1 - p_i$;
and $|\mathcal{L}|$ is the number of models.

\subsection{Detailed Explanation of the Common-Person Calibration Procedure}
\label{common_person_calibration}

In this work, calibration refers to estimating item characteristics under the 3PL item response theory model. The purpose of calibration is to place all items on a shared difficulty scale so that performance comparisons across items and models become meaningful. Under the 3PL model, each item is described by a difficulty parameter, a discrimination parameter that captures how strongly the item differentiates between high- and low-performing models, and a guessing parameter that reflects the chance of a correct response when the model effectively guesses. When every LLM responds to the same items, their collective performance patterns allow these parameters to be estimated in a consistent way. This provides a principled way to identify which items are easy or difficult for models and which items are more or less informative.
Calibrating the full benchmark at once would be computationally intensive because the 3PL model becomes more expensive to fit as the number of items grows. To make the process tractable, we divide the full item pool into several non-overlapping subsets and calibrate each subset independently. This reduces the computational load substantially, but it also means that each subset is estimated on its own internal scale. For example, the notion of “difficulty” in one subset is not automatically aligned with the notion of “difficulty” in another. A separate linking step is therefore required to place all subsets onto a shared scale.

Linking requires shared reference points known as anchors. In educational measurement, anchors are typically common items or common examinees that appear in multiple test forms. They serve as a bridge that allows independently calibrated scales to be aligned. In our setting, every LLM responds to every item in the benchmark, which means that the same population of models appears in the calibration of each item subset. The models therefore act as common persons in the traditional psychometric sense and serve as the linking anchors for the benchmark. Their relative performance across subsets provides the information needed to align the scale of each subset with the others.
The linking procedure examines how the same models perform across the different subsets and adjusts each subset’s scale so that the overall performance patterns match. If a model appears stronger than its peers in one subset and shows a similar relative standing in another, then the two subsets can be placed on the same scale by aligning the average performance level and the overall spread of performance. This rescaling is then applied to the item parameters of each subset so that all items, regardless of which subset they came from, are expressed on a single, unified difficulty metric.

This form of common-person linking is particularly effective for LLM benchmarking. In human testing, it is rarely feasible for every examinee to respond to every item, and linking must rely on smaller or less reliable anchor sets. LLMs do not face constraints such as fatigue, practice effects, or time limits, which allows us to use the entire model population as a complete and stable set of anchors. This makes the linking process highly robust and enables a scalable calibration framework that achieves substantial computational efficiency while maintaining coherence across a very large item bank.

\begin{table*}[ht]
\centering

\renewcommand{\arraystretch}{1.2}
\begin{tabular}{lccccc}
\hline
 & \textbf{WinoGrande} & \textbf{TruthfulQA} & \textbf{HellaSwag} & \textbf{GSM8K} & \textbf{ARC} \\
\hline
\textbf{\# Models used for calibration}       & 4680 & 4635 & 3467 & 3775 & 3747 \\
\textbf{\# Models used for testing}           & 521  & 516  & 386  & 420  & 417  \\
\textbf{\# Calibration subsets ($K$)}         & 10   & 6    & 50   & 12   & 8    \\
\textbf{\# Items after filtering}             & 1045 & 627  & 5600 & 1306 & 839  \\
\textbf{Average RMSEA ($M_2$)}             & 0.0565 & 0.0690 & 0.0482 & 0.0438 & 0.0595 \\
\hline
\end{tabular}
\caption{
Statistics describing the calibration dataset, testing dataset, item-bank size after filtering, number of calibration partitions (K), and average model–data fit (RMSEA from M2) across all benchmarks.
}
\label{tab:filtering_summary}
\end{table*}

\subsection{Calibration Data, Item-Bank Partitioning, and Fit Statistics}

A summary of the calibration data, item-bank partitioning, and associated fit statistics is provided in Table~\ref{tab:filtering_summary}.

Beyond the M2/RMSEA statistics, we additionally evaluate local independence using Yen's Q3 residual correlations \citep{yen1984effects}. Table~\ref{tab:q3} reports mean Q3 values, adjusted Q3 percentiles \citep{christensen2017critical}, and the proportion of item pairs exceeding commonly used dependence thresholds \citep{efd59699adde42568af11fe1ed801e45}.

Across all benchmarks, mean off-diagonal Q3 values remain close to zero, suggesting that local independence holds reasonably well at the benchmark level. For ARC and HellaSwag, the 99th-percentile adjusted Q3 values are approximately 0.20, with only around 1\% of item pairs exceeding the 0.20 threshold. These results indicate that elevated residual dependence exists only for a small subset of item pairs, likely due to shared templates or topical overlap, rather than broad violations of local independence.
\begin{table}[t]
\centering
\small
\setlength{\tabcolsep}{5pt}
\begin{tabular}{lcccc}
\toprule
\textbf{Benchmark} &
\textbf{Mean Q3} &
\textbf{P99 adj.\ Q3} &
\textbf{Prop.\ adj.\ Q3 $>$ 0.20} &
\textbf{Prop.\ adj.\ Q3 $>$ 0.30} \\
\midrule

HellaSwag & 0.004 & 0.202 & 0.010 & 0.002 \\
GSM8K & 0.002 & 0.144 & 0.002 & 0.000 \\
TruthfulQA & 0.011 & 0.257 & 0.021 & 0.006 \\
ARC & 0.011 & 0.208 & 0.011 & 0.002 \\
WinoGrande & 0.007 & 0.187 & 0.008 & 0.001 \\

\bottomrule
\end{tabular}
\caption{
Yen's Q3 residual correlation statistics for evaluating local independence.
Mean off-diagonal Q3 values remain close to zero across benchmarks,
while only a small proportion of item pairs exceed common adjusted-Q3 dependence thresholds.
}
\label{tab:q3}
\end{table}

\section{Detailed Experimental Setup and Metrics}
\label{app:experimental_details}

\subsection{Benchmarks and Datasets}
We conduct experiments on five diverse benchmarks covering different cognitive domains:
\begin{itemize}
    \item \textbf{WinoGrande}: Commonsense reasoning with pronoun resolution
    \item \textbf{TruthfulQA}: Factual consistency and truthfulness evaluation
    \item \textbf{HellaSwag}: Procedural inference and common sense completion
    \item \textbf{GSM8K}: Mathematical word problems requiring multi-step reasoning
    \item \textbf{ARC}: Scientific question answering across multiple domains
\end{itemize}

All experiments use the calibrated item banks from Section~\ref{sec:calibration}, ensuring consistent filtering and parameter quality across datasets. 

\subsection{Baseline Configurations}
We compare \CAT\ against four fixed, non-adaptive strategies:

\textbf{Random Baseline:} Samples 100 items uniformly from the full bank without consideration of item parameters or model ability.

\textbf{TinyBenchmarks:} Uses the predetermined subset from \cite{polo2024tinybenchmarks}, selected via clustering methods but without explicit Fisher information optimization for ability estimation.

\textbf{MetaBench-Primary and MetaBench-Secondary:} Curated splits from \cite{metabench} that require computationally expensive iterations to identify stable subsets. These splits emphasize predictive accuracy over psychometric validity.

All baseline data is available on Hugging Face: \href{https://huggingface.co/tinyBenchmarks}{tinyBenchmarks}, \href{https://huggingface.co/datasets/HCAI/metabench}{HCAI/metabench}.

Unlike \CAT, these approaches do not adapt to individual test-takers and serve only as static reference points for accuracy–efficiency tradeoffs.

\subsection{\CAT\ Configuration Details}
For each model $\ell$, we run \CAT\ under three precision-based stopping thresholds:
\begin{itemize}
    \item $\mathrm{SE}(\hat{\theta}) \leq 0.1$: High precision, suitable for fine-grained model comparison
    \item $\mathrm{SE}(\hat{\theta}) \leq 0.2$: Moderate precision, balancing accuracy and efficiency
    \item $\mathrm{SE}(\hat{\theta}) \leq 0.3$: Lower precision, maximizing efficiency for rapid screening
\end{itemize}

A minimum of 30 items is enforced to prevent premature termination due to lucky guesses or initial high-information items, while the maximum is capped at 500 items to ensure computational feasibility. This setup balances precision and budget constraints, simulating realistic conditions for adaptive evaluation in production environments.

\section{Evaluation Metric Definitions}
\label{app:metrics}

This section provides the exact mathematical definitions of the evaluation metrics introduced in Section~\ref{sec:results}, along with brief interpretations.

\paragraph{Average Mean Absolute Error (MAE) and Standard Error (SE).}
We compute MAE for both ability estimates and accuracy scores. For ability, let
$\hat{\theta}_\ell$ denote the CAT-derived estimate and
$\hat{\theta}_\ell^{\mathrm{whole}}$ the full-bank reference. The ability MAE is
\[
\text{MAE}_{\theta}
= \frac{1}{|\mathcal{L}|} \sum_{\ell \in \mathcal{L}}
\left| \hat{\theta}_\ell - \hat{\theta}_\ell^{\mathrm{whole}} \right|.
\]
For accuracy, let $\widehat{Acc}^{\mathrm{p\text{-}IRT}}_\ell$ denote the reconstructed accuracy
(e.g., via p-IRT) and $\mathrm{Acc}_\ell^{\mathrm{raw}}$ the observed raw accuracy.
The accuracy MAE is
\[
\text{MAE}_{\mathrm{acc}}
= \frac{1}{|\mathcal{L}|} \sum_{\ell \in \mathcal{L}}
\left| \widehat{Acc}^{\mathrm{p\text{-}IRT}}_\ell - \mathrm{Acc}_\ell^{\mathrm{raw}} \right|.
\]

To quantify variability across models, we also report the standard error (SE) of
MAE. Let $e_\ell$ denote the per-model absolute error and $\overline{e}$ its mean.
The standard deviation (SD) is
\[
\text{SD}
= \sqrt{ \frac{1}{|\mathcal{L}| - 1}
         \sum_{\ell \in \mathcal{L}}
         (e_\ell - \overline{e})^{2} },
\]
and the standard error (SE) is
\[
\text{SE} = \frac{\text{SD}}{\sqrt{|\mathcal{L}|}}.
\]

\textit{Interpretation:} Lower MAE and SE indicate higher fidelity and greater
stability: CAT-derived estimates more closely match whole-bank references (for
ability) or observed scores (for accuracy) and do so consistently across models.

\paragraph{Information Efficiency Score (IES).}
To compare the efficiency of different evaluation methods, we define the
\textit{Information Efficiency Score} (IES) relative to a baseline of 100-item
uniform random sampling (Random\_100). For a given method, let
$\text{MAE}_{\text{method}}$ denote its average MAE and $\text{MAE}_{\text{Random}}$
the MAE under Random\_100. Let $\text{Items}_{\text{method}}$ denote the average
number of selected subset items. The IES is:

\begin{equation}
    \text{IES} \;=\;
    \left( \frac{\text{MAE}_{\text{method}}}{\text{MAE}_{\text{Random}}} \right)
    \left( \frac{\text{Items}_{\text{method}}}{100} \right).
\end{equation}

\textit{Interpretation:}  
An IES value below 1 indicates that the method achieves a better
accuracy–efficiency tradeoff than the Random\_100 baseline, requiring fewer items
and/or producing lower error for the same number of items. An IES value of 1
means the method is equally efficient as Random\_100. Values greater than 1
indicate lower efficiency, meaning the method uses more items and/or yields
higher error than the baseline.

\paragraph{Average Item Exposure Rate.}  
Let $h_i$ denote the number of models administered item $i$, with $|\mathcal{I}|$ total items and $|\mathcal{L}|$ total models. The item exposure probability for item $i$ is
\begin{equation}
    P(A_i) = \frac{h_i}{|\mathcal{L}|}.
\end{equation}
The average item exposure rate is then
\begin{equation}
    \bar{P}(A_i) = \frac{1}{|\mathcal{I}|} \sum_{i \in \mathcal{I}} P(A_i).
\end{equation}  
\textit{Interpretation:} Lower values indicate higher adaptivity and greater item diversity, while higher values suggest uniform or repetitive item usage across models.

\paragraph{Test Overlap Rate.}  
Following \citet{chen2005simplified}, the expected proportion of common items between two randomly selected test forms is given by
\begin{equation}
    \bar{Q} = \frac{|\mathcal{L}| \sum_{i=1}^{|\mathcal{I}|} P(A_i)^2}{\bar{L}(|\mathcal{L}| - 1)} - \frac{1}{|\mathcal{L}| - 1},
\end{equation}
where $\bar{L}$ is the average test length.  

\textit{Interpretation:} Lower values of $\bar{Q}$ imply greater test form diversity, which reduces risks of collusion and item memorization.

\subsection{Correlation Metrics}
\label{app:correlation}

For completeness, we provide the definitions of the rank-based correlation coefficients used in Section~\ref{sec:results}.

\paragraph{Spearman correlation.}  
\[
\rho = 1 - \frac{6 \sum_{i=1}^n d_i^2}{n(n^2-1)},
\]
where $d_i$ is the rank difference for observation $i$ across the two measures.

\paragraph{Kendall correlation.}  
\[
\tau = \frac{(\#\text{concordant pairs}) - (\#\text{discordant pairs})}{\tfrac{1}{2}n(n-1)}.
\]

\section{Performance-IRT (p-IRT) Estimator} 
\label{app:p-irt}

The Performance-IRT (p-IRT) estimator \citep{polo2024tinybenchmarks} is a probabilistic scoring method used to
compute expected accuracy when only a subset of benchmark items is observed.
Conceptually, p-IRT is grounded in the Test Characteristic Curve (TCC)
\citep{lord2008statistical, hambleton1991fundamentals}, which maps a model’s
ability $\hat{\theta}$ to its expected probability of correctly answering items
under the calibrated 3PL model. It provides an estimate of a model’s overall
benchmark accuracy without requiring evaluation on the full item set.

\paragraph{Goal.}
We formulate LLM evaluation as a psychometric measurement problem. Let
$\mathcal{I}$ denote the full set of benchmark items and $\mathcal{L}$ the set
of language models. For each model $\ell \in \mathcal{L}$ and item
$i \in \mathcal{I}$, we observe a binary response
$Y_{i,\ell} \in \{0,1\}$, forming the item–response matrix
$\{Y_{i,\ell}\}_{i \in \mathcal{I},\,\ell \in \mathcal{L}}$.
The true full-benchmark accuracy of model $\ell$ is
\[
\mathrm{Acc}_\ell^{\mathrm{raw}} = \frac{1}{|\mathcal{I}|}\sum_{i \in \mathcal{I}} Y_{i,\ell}.
\]
The p-IRT estimator approximates $\mathrm{Acc}_\ell^{\mathrm{raw}}$ when only a subset of items
$\widehat{\mathcal{I}} \subseteq \mathcal{I}$ is observed, by combining the
model’s observed responses on $\widehat{\mathcal{I}}$ with IRT-predicted
probabilities on the remaining items $\mathcal{I} \setminus \widehat{\mathcal{I}}$.

\paragraph{Estimator.}
P-IRT computes the conditional expectation
\[
\widehat{Acc}^{\mathrm{p\text{-}IRT}}_\ell
    = \mathbb{E}\!\left[ \mathrm{Acc}_\ell^{\mathrm{raw}}
      \,\middle|\,
      \{Y_{i,\ell} : i \in \widehat{\mathcal{I}}\}
      \right],
\]
which is the minimum–mean-squared-error predictor of $\mathrm{Acc}_\ell^{\mathrm{raw}}$ under the
calibrated IRT model. Then
\[
\widehat{Acc}^{\mathrm{p\text{-}IRT}}_\ell
=
\frac{|\widehat{\mathcal{I}}|}{|\mathcal{I}|}
\cdot \underbrace{\frac{1}{|\widehat{\mathcal{I}}|}
\sum_{i \in \widehat{\mathcal{I}}} Y_{i,\ell}
}_{\text{Observed accuracy}}
\;+\;
\frac{|\mathcal{I} \setminus \widehat{\mathcal{I}}|}{|\mathcal{I}|}
\cdot
\underbrace{\frac{1}{|\mathcal{I} \setminus \widehat{\mathcal{I}}|}
\sum_{i \in \mathcal{I} \setminus \widehat{\mathcal{I}}}
\hat{p}_{i,\ell}
}_{\text{Unobserved TCC}},
\]
where
\[
\hat{p}_{i,\ell}
    = P\!\left(Y_{i,\ell}=1 \mid
        \hat{\theta}_\ell, \hat{a}_i, \hat{b}_i, \hat{c}_i
      \right)
\]
is the predicted probability of correctness for model $\ell$ under the
calibrated 3PL model.

\paragraph{Intuition.}
The p-IRT estimator is a weighted combination of:
\begin{itemize}
    \item \textbf{Observed accuracy} on the subset $\widehat{\mathcal{I}}$.
    \item \textbf{Unobserved TCC} on the remaining items
          $\mathcal{I} \setminus \widehat{\mathcal{I}}$, based on the model’s
          ability $\hat{\theta}_\ell$ and item parameters.
\end{itemize}
The weight $|\widehat{\mathcal{I}}| / |\mathcal{I}| \in [0,1]$ corresponds to the proportion of observed
items and determines the tradeoff between observed and predicted performance.

\paragraph{Use in This Work.}
We apply p-IRT to reconstruct accuracy from ability estimates
$\hat{\theta}_\ell$. 
As shown in Table~\ref{tab:acc_results}, reconstructed accuracies closely match
raw accuracies across benchmarks, confirming that ability estimates retain the
global performance structure while smoothing noise and offering finer
discrimination than raw accuracy alone.

\section{Additional Experimental Results}
\label{app:additional_results}

\subsection{Additional Benchmark Results}
\label{app:additional_tables}

Tables~\ref{tab:main_result_appendix} and~\ref{tab:acc_results_appendix} present results for GSM8K and ARC benchmarks, complementing the main results in Tables~\ref{tab:main_result} and~\ref{tab:acc_results}.

\begin{table*}[ht]
\centering
\small

\begin{tabular}{l|ccc|ccc}
\toprule
\multirow{2}{*}{\textbf{Method}} &
\multicolumn{3}{c|}{\textbf{GSM8K}} &
\multicolumn{3}{c}{\textbf{ARC}} \\
& MAE$\pm$SE $\downarrow$  & Items $\downarrow$  & IES $\downarrow$
& MAE$\pm$SE $\downarrow$  & Items $\downarrow$  & IES $\downarrow$ \\
\midrule

Random$_{100}$
& \dashuline{0.150$\pm$0.014} & 100 & 1.000
& 0.183$\pm$0.007 & 100 & 1.000 \\

TinyBenchmarks
& 0.164$\pm$0.014 & 100 & 1.089
& 0.172$\pm$0.007 & 99 & 0.932 \\

MetaBench-P
& \underline{0.103$\pm$0.013} & 237 & 1.628
& 0.134$\pm$0.005 & 145 & 1.062 \\

MetaBench-S
& \textbf{0.096$\pm$0.012} & 249 & 1.595
& 0.134$\pm$0.006 & 100 & 0.735 \\

ATLAS$_{0.1}$
& \dashuline{0.150$\pm$0.011} & \dashuline{70} & \dashuline{0.701}
& \textbf{0.084$\pm$0.006} & \dashuline{89} & \dashuline{0.407} \\

ATLAS$_{0.2}$
& 0.177$\pm$0.012 & \underline{36} & \underline{0.428}
& \dashuline{0.120$\pm$0.008} & \underline{35} & \underline{0.232} \\

ATLAS$_{0.3}$
& 0.173$\pm$0.012 & \textbf{31} & \textbf{0.363}
& \underline{0.117$\pm$0.007} & \textbf{30} & \textbf{0.193} \\

\bottomrule
\end{tabular}
\caption{Comparison of whole-bank ability $\hat{\theta}_{\ell}^{\text{whole}}$ and subset-based
ability $\hat{\theta}_{\ell}$ on GSM8K and ARC benchmarks. For each method, we report MAE$\pm$SE,
item count, and Information Efficiency Score (IES), where lower values are
better for all metrics.}
\label{tab:main_result_appendix}
\end{table*}

\begin{table*}[ht]
\centering
\small

\begin{tabular}{l|ccc|ccc}
\toprule
\multirow{2}{*}{\textbf{Method}} &
\multicolumn{3}{c|}{\textbf{GSM8K}} &
\multicolumn{3}{c}{\textbf{ARC}} \\
& MAE$\pm$SE $\downarrow$  & Items $\downarrow$  & IES $\downarrow$
& MAE$\pm$SE $\downarrow$  & Items $\downarrow$  & IES $\downarrow$ \\
\midrule

Random$_{100}$
& \dashuline{0.026$\pm$0.001} & 100 & \dashuline{1.000}
& \underline{0.029$\pm$0.001} & 100 & 1.000 \\

TinyBenchmarks
& 0.028$\pm$0.001 & 100 & 1.071
& \dashuline{0.031$\pm$0.001} & 99 & 1.041 \\

MetaBench-P
& \underline{0.022$\pm$0.001} & 237 & 2.060
& \textbf{0.027$\pm$0.001} & 145 & 1.350 \\

MetaBench-S
& \textbf{0.020$\pm$0.001} & 249 & 1.954
& 0.033$\pm$0.001 & 100 & 1.114 \\

ATLAS$_{0.1}$
& 0.039$\pm$0.001 & \dashuline{70} & 1.055
& 0.032$\pm$0.002 & \dashuline{89} & \dashuline{0.974} \\

ATLAS$_{0.2}$
& 0.044$\pm$0.002 & \underline{36} & \underline{0.612}
& 0.034$\pm$0.002 & \underline{35} & \underline{0.404} \\

ATLAS$_{0.3}$
& 0.042$\pm$0.002 & \textbf{31} & \textbf{0.516}
& 0.034$\pm$0.002 & \textbf{30} & \textbf{0.350} \\

\bottomrule
\end{tabular}
\caption{Comparison of raw whole-bank accuracy and p-IRT reconstructed accuracy on GSM8K and ARC benchmarks.
For each method, we report MAE$\pm$SE, item count, and the
Information Efficiency Score (IES), where lower values are better for all metrics.}
\label{tab:acc_results_appendix}
\end{table*}

 \subsection{Interpreting Test Overlap, Exposure, and Runtime in ATLAS}
\label{app:table6_analysis}

Table~\ref{tab:results_test_overlap} provides a detailed breakdown of adaptive
evaluation behavior across all five benchmarks, summarizing test overlap,
average item exposure, and selection time. These metrics together illustrate how
ATLAS balances efficiency, diversity, and computational scalability when
administering adaptive tests. Formal definitions of all metrics are included in Appendix~\ref{app:metrics}.

Test overlap rates quantify how frequently different models are exposed to the
same items. Across benchmarks, overlap remains modest, ranging from roughly
11\% to 23\%. These values are far lower than those of static subsets, which
administer identical items to all models. The relatively low overlap indicates
that ATLAS tailors item sequences to each model's evolving ability estimate
rather than relying on a fixed set of questions. For example, HellaSwag reaches
the lowest overlap values (as low as 11.26\%), reflecting its large item pool
and the wide range of informative items available. Higher overlap on datasets
such as GSM8K (approximately 20--24\%) reflects the smaller bank size and the
concentration of discriminative items in particular ability regions. Overall,
the overlap statistics confirm that ATLAS provides genuine adaptivity while
preserving comparability across models.

Average item exposure rates remain low across all settings, consistently under
12\% and often much lower. Exposure values around 3--5\% on HellaSwag and
WinoGrande (for SE thresholds of 0.2 and 0.3) indicate that ATLAS does not rely
excessively on a small subset of items. Low exposure reduces the risk of
memorization or contamination in long-term evaluation scenarios, broadens the
portion of the item bank that contributes to measurement, and ensures that no
individual item disproportionately influences ability estimation. The pattern
across SE thresholds reflects a standard property of adaptive testing: when
fewer items are required (larger SE thresholds), exposure becomes more
concentrated on the most informative items. In ATLAS, this concentration
remains moderate, indicating healthy rotation among informative items.

Selection time reflects the computational cost of the full adaptive selection
loop, including Fisher information computation and termination checks. It refers to the complete runtime of the adaptive item selection loop for each model, rather than the time required for a single item decision. Times
range from 9 to 76 seconds per model and scale predictably with benchmark size.
TruthfulQA, with 628 items, achieves the fastest selection times (approximately
9--16 seconds). In contrast, HellaSwag, with more than 5600 items, shows the
longest selection times (57--76 seconds), due to the larger number of items
evaluated when determining the most informative question at each step.
Importantly, even in the largest setting, selection remains well under 90
seconds, and for all other benchmarks it typically completes within tens of
seconds. This confirms that ATLAS is computationally practical for both
interactive evaluation and large-scale benchmarking workflows.

Taken together, the results in Table~\ref{tab:results_test_overlap} show that
ATLAS achieves high adaptivity, broad item utilization, and practical runtime
efficiency across diverse benchmarks. Low overlap and exposure promote content
coverage and robustness, while stable runtime performance ensures operational
scalability without compromising statistical quality.

\begin{table*}[ht]
\centering

\begin{tabular}{c|c|c|c|c}
\toprule
\textbf{Benchmark} & \textbf{Method} & \textbf{Test Overlap $\downarrow$ } & \textbf{Avg. Item $\downarrow$ } & \textbf{Avg. Selection $\downarrow$} \\
 (Item \#)&  & \textbf{Rate} (\%) & \textbf{Exposure Rate} (\%) & \textbf{Time} (s)   \\
\midrule
\multirow{3}{*}{\makecell[c]{WinoGrande\\(1046)}}
& ATLAS$_{SE \leq 0.1}$ & \underline{18.22} & \dashuline{8.24} & \dashuline{40.99} \\
& ATLAS$_{SE \leq 0.2}$ & \textbf{14.93} & \underline{4.71} & \underline{19.92} \\
& ATLAS$_{SE \leq 0.3}$ & \dashuline{17.03} & \textbf{4.04} & \textbf{16.74} \\
\midrule
\multirow{3}{*}{\makecell[c]{TruthfulQA\\(628)}}
& ATLAS$_{SE \leq 0.1}$ & \textbf{17.32} & \textbf{7.86} & \dashuline{15.97} \\
& ATLAS$_{SE \leq 0.2}$ & \dashuline{18.43} & \underline{9.58} & \textbf{9.37} \\
& ATLAS$_{SE \leq 0.3}$ & \underline{18.07} & \dashuline{9.49} & \underline{9.72} \\
\midrule
\multirow{3}{*}{\makecell[c]{HellaSwag\\(5600)}}
& ATLAS$_{SE \leq 0.1}$ & \textbf{11.26} & \textbf{3.86} & \dashuline{75.52} \\
& ATLAS$_{SE \leq 0.2}$ & \dashuline{13.72} & \underline{4.78} & \textbf{56.93} \\
& ATLAS$_{SE \leq 0.3}$ & \underline{13.67} & \dashuline{4.82} & \underline{57.06} \\
\midrule
\multirow{3}{*}{\makecell[c]{GSM8K\\(1307)}}
& ATLAS$_{SE \leq 0.1}$ & \underline{21.27} & \dashuline{7.21} & \dashuline{45.69} \\
& ATLAS$_{SE \leq 0.2}$ & \textbf{20.78} & \textbf{4.40} & \underline{24.08} \\
& ATLAS$_{SE \leq 0.3}$ & \dashuline{23.70} & \underline{5.54} & \textbf{19.06} \\
\midrule
\multirow{3}{*}{\makecell[c]{ARC\\(842)}}
& ATLAS$_{SE \leq 0.1}$ & \underline{19.15} & \dashuline{11.15} & \dashuline{30.99} \\
& ATLAS$_{SE \leq 0.2}$ & \textbf{17.09} & \textbf{5.41} & \underline{13.98} \\
& ATLAS$_{SE \leq 0.3}$ & \dashuline{19.60} & \underline{9.21} & \textbf{11.82} \\
\bottomrule
\end{tabular}
\caption{Adaptive evaluation efficiency and diversity. \CAT\ maintains low item exposure rates ($<12\%$) and moderate test overlap ($13-24\%$) with fast selection times ($<76$ seconds per model). Lower values indicate better performance for all metrics.}
\label{tab:results_test_overlap}
\end{table*}

\subsection{Sensitivity to the IRT Parameterization}
\label{app:2pl3pl}

We compare the default 3PL parameterization with a 2PL variant
that removes the lower-asymptote parameter $c_i$. In human
testing, $c_i$ is often interpreted as a guessing parameter. In the
LLM setting, however, this interpretation should be treated with
caution: the lower asymptote may reflect answer priors, prompting
conventions, decoding behavior, memorization, or learned
heuristics, rather than literal random guessing.

Table~\ref{tab:2pl3pl_combined}  reports
the comparison for ability estimation and raw-accuracy
reconstruction. The result shows that 3PL is not
uniformly superior across all settings. On TruthfulQA, 3PL
consistently outperforms 2PL under all stopping thresholds for
both ability and accuracy reconstruction. On the remaining
benchmarks, the comparison is mixed: 2PL sometimes achieves
lower ability MAE, especially under stricter stopping thresholds,
but often requires more items. In contrast, 3PL remains competitive
in ability estimation and frequently provides better accuracy
reconstruction, particularly on ARC and HellaSwag. These results
indicate that 3PL should be viewed as a flexible modeling choice
rather than a universally required parameterization.
\begin{table*}
\centering
\small
\setlength{\tabcolsep}{4pt}


\begin{tabular}{llcccccccc}
\toprule
& &
\multicolumn{4}{c}{\textbf{Ability Estimation}} &
\multicolumn{4}{c}{\textbf{Accuracy Reconstruction}} \\
\cmidrule(lr){3-6}
\cmidrule(lr){7-10}
\textbf{Benchmark} & \textbf{SE} &\multicolumn{2}{c}{\textbf{3PL}} &
\multicolumn{2}{c}{\textbf{2PL}} &
\multicolumn{2}{c}{\textbf{3PL}} &
\multicolumn{2}{c}{\textbf{2PL}} \\
& & MAE$\pm$SE $\downarrow$  & Items $\downarrow$ & MAE$\pm$SE $\downarrow$  & Items $\downarrow$ & MAE$\pm$SE $\downarrow$  & Items $\downarrow$ & MAE$\pm$SE $\downarrow$  & Items $\downarrow$ \\

\midrule

WinoGrande & 0.1
& 0.155 $\pm$ 0.012 & 70
& \textbf{0.121 $\pm$ 0.010} & 198
& 0.048 $\pm$ 0.001 & 70
& \textbf{0.025 $\pm$ 0.001} & 198 \\

WinoGrande & 0.2
& \textbf{0.166 $\pm$ 0.010} & 37
& 0.170 $\pm$ 0.010 & 77
& 0.051 $\pm$ 0.002 & 37
& \textbf{0.034 $\pm$ 0.001} & 77 \\

WinoGrande & 0.3
& \textbf{0.179 $\pm$ 0.011} & 32
& 0.197 $\pm$ 0.011 & 51
& 0.050 $\pm$ 0.001 & 32
& \textbf{0.036 $\pm$ 0.001} & 51 \\

\midrule

TruthfulQA & 0.1
& \textbf{0.064 $\pm$ 0.002} & 48
& 0.078 $\pm$ 0.004 & 61
& \textbf{0.023 $\pm$ 0.001} & 48
& 0.049 $\pm$ 0.001 & 61 \\

TruthfulQA & 0.2
& \textbf{0.073 $\pm$ 0.003} & 30
& 0.095 $\pm$ 0.006 & 30
& \textbf{0.024 $\pm$ 0.001} & 30
& 0.052 $\pm$ 0.001 & 30 \\

TruthfulQA & 0.3
& \textbf{0.071 $\pm$ 0.003} & 30
& 0.097 $\pm$ 0.006 & 30
& \textbf{0.023 $\pm$ 0.001} & 30
& 0.052 $\pm$ 0.001 & 30 \\

\midrule

GSM8K & 0.1
& 0.150 $\pm$ 0.011 & 70
& \textbf{0.132 $\pm$ 0.007} & 100
& 0.039 $\pm$ 0.001 & 70
& \textbf{0.036 $\pm$ 0.001} & 100 \\

GSM8K & 0.2
& 0.177 $\pm$ 0.012 & 36
& \textbf{0.172 $\pm$ 0.008} & 35
& 0.044 $\pm$ 0.002 & 36
& \textbf{0.042 $\pm$ 0.002} & 35 \\

GSM8K & 0.3
& \textbf{0.173 $\pm$ 0.012} & 31
& 0.175 $\pm$ 0.008 & 31
& \textbf{0.042 $\pm$ 0.002} & 31
& 0.042 $\pm$ 0.002 & 31 \\

\midrule

ARC & 0.1
& 0.084 $\pm$ 0.006 & 89
& \textbf{0.076 $\pm$ 0.004} & 135
& \textbf{0.032 $\pm$ 0.002} & 89
& 0.068 $\pm$ 0.002 & 135 \\

ARC & 0.2
& 0.120 $\pm$ 0.008 & 35
& \textbf{0.107 $\pm$ 0.005} & 64
& \textbf{0.034 $\pm$ 0.002} & 35
& 0.073 $\pm$ 0.002 & 64 \\

ARC & 0.3
& 0.117 $\pm$ 0.007 & 30
& \textbf{0.116 $\pm$ 0.006} & 33
& \textbf{0.034 $\pm$ 0.002} & 30
& 0.077 $\pm$ 0.002 & 33 \\

\midrule

HellaSwag & 0.1
& 0.157 $\pm$ 0.010 & 41
& \textbf{0.141 $\pm$ 0.006} & 81
& \textbf{0.020 $\pm$ 0.001} & 41
& 0.035 $\pm$ 0.001 & 81 \\

HellaSwag & 0.2
& \textbf{0.163 $\pm$ 0.009} & 30
& 0.168 $\pm$ 0.008 & 32
& \textbf{0.021 $\pm$ 0.001} & 30
& 0.035 $\pm$ 0.001 & 32 \\

HellaSwag & 0.3
& \textbf{0.165 $\pm$ 0.010} & 30
& 0.170 $\pm$ 0.008 & 30
& \textbf{0.021 $\pm$ 0.001} & 30
& 0.035 $\pm$ 0.001 & 30 \\

\bottomrule
\end{tabular}

\caption{
Comparison between 3PL and 2PL parameterizations for both ability estimation and raw-accuracy reconstruction. Columns denote ATLAS stopping thresholds (SE $\leq$ 0.1/0.2/0.3). Lower MAE is better.
}

\label{tab:2pl3pl_combined}
\end{table*}

\subsection{Sensitivity to the Ability Estimator}

Our main experiments use Weighted Likelihood Estimation (WLE) for whole-bank reference ability estimation and Expected A Posteriori (EAP) estimation during ATLAS inference. WLE provides stable finite estimates across a broad ability range, while EAP is numerically more stable during early adaptive evaluation with sparse responses.

To evaluate sensitivity to this estimator choice, we additionally compute whole-bank reference abilities using EAP instead of WLE. Table~\ref{tab:wle_eap} compares ATLAS ability estimation MAE under both reference estimators across five representative benchmarks. Overall, the estimator choice affects some benchmarks more than others, but the main conclusion remains unchanged: ATLAS closely tracks whole-bank ability estimates while using substantially fewer evaluation items under both WLE- and EAP-based references.
\begin{table*}[t]
\centering
\small
\setlength{\tabcolsep}{5pt}
\begin{tabular}{lcccccc}
\toprule
& \multicolumn{2}{c}{\textbf{ATLAS$_{SE \leq 0.1}$}} 
& \multicolumn{2}{c}{\textbf{ATLAS$_{SE \leq 0.2}$}} 
& \multicolumn{2}{c}{\textbf{ATLAS$_{SE \leq 0.3}$}} \\
\cmidrule(lr){2-3}
\cmidrule(lr){4-5}
\cmidrule(lr){6-7}

\textbf{Benchmark}
& \textbf{WLE}
& \textbf{EAP}
& \textbf{WLE}
& \textbf{EAP}
& \textbf{WLE}
& \textbf{EAP} \\

& MAE$\pm$SE $\downarrow$ 
& MAE$\pm$SE $\downarrow$ 
& MAE$\pm$SE $\downarrow$ 
& MAE$\pm$SE $\downarrow$ 
& MAE$\pm$SE $\downarrow$ 
& MAE$\pm$SE $\downarrow$ 
\\
\midrule
WinoGrande
& 0.155 $\pm$ 0.012
& \textbf{0.136 $\pm$ 0.011}
& \textbf{0.166 $\pm$ 0.010}
& 0.167 $\pm$ 0.013
& \textbf{0.179 $\pm$ 0.011}
& 0.184 $\pm$ 0.014 \\

TruthfulQA
& \textbf{0.064 $\pm$ 0.002}
& 0.072 $\pm$ 0.005
& \textbf{0.073 $\pm$ 0.003}
& 0.083 $\pm$ 0.007
& \textbf{0.071 $\pm$ 0.003}
& 0.074 $\pm$ 0.005 \\

HellaSwag
& 0.157 $\pm$ 0.010
& \textbf{0.087 $\pm$ 0.005}
& 0.163 $\pm$ 0.009
& \textbf{0.099 $\pm$ 0.006}
& 0.165 $\pm$ 0.010
& \textbf{0.093 $\pm$ 0.006} \\

GSM8K
& 0.150 $\pm$ 0.011
& \textbf{0.138 $\pm$ 0.009}
& 0.177 $\pm$ 0.012
& \textbf{0.153 $\pm$ 0.010}
& 0.173 $\pm$ 0.012
& \textbf{0.164 $\pm$ 0.011} \\

ARC
& \textbf{0.084 $\pm$ 0.006}
& 0.089 $\pm$ 0.007
& 0.120 $\pm$ 0.008
& \textbf{0.116 $\pm$ 0.009}
& \textbf{0.117 $\pm$ 0.007}
& 0.126 $\pm$ 0.010 \\

\bottomrule
\end{tabular}
\caption{
Comparison of ATLAS ability estimation MAE using WLE- and EAP-based whole-bank reference abilities under different stopping thresholds. Lower MAE indicates better agreement with whole-bank estimates.
}
\label{tab:wle_eap}
\end{table*}

\subsection{Experiment on MMLU}
While both \textit{TinyBenchmarks} \citep{polo2024tinybenchmarks} and \textit{MetaBench} \citep{metabench} include MMLU \citep{hendrycks2020measuring} as part of their evaluation suites, they treat it as a single unified dataset by aggregating all 57 subject areas. In contrast, we perform evaluation on a per-subject basis. This design choice acknowledges the heterogeneous nature of MMLU, where each subject represents a distinct knowledge domain with varying linguistic characteristics, content distributions, and difficulty levels. Aggregating across all subjects can obscure these domain-specific patterns and limit interpretability in adaptive assessment.

The corresponding results are reported in Table~\ref{tab:mmlu_adaptive_results}. Despite the small number of items per subject, \textsc{ATLAS} consistently demonstrates robust adaptive evaluation performance. As the selection threshold is relaxed ($SE \leq 0.1 \rightarrow 0.3$), the mean absolute error (MAE) increases moderately (e.g., from 0.099 to 0.235 in \textit{Anatomy}), while the number of evaluated items is substantially reduced (approximately 80\%). This indicates that \textsc{ATLAS} effectively balances efficiency and accuracy, even in limited-data regimes.

Moreover, reductions in test overlap and exposure rates across the evaluated subjects suggest that the adaptive mechanism achieves broader item coverage and mitigates redundancy. Evaluation time also decreases proportionally with the number of items, confirming the computational efficiency of the adaptive process.

\begin{scriptsize} 
\setlength{\tabcolsep}{3pt} 

\begin{longtable}{c|c|c|c|c|c|c}
\caption{Adaptive evaluation results on MMLU subsets (item range: 10–100). Lower is better $\downarrow$.}
\label{tab:mmlu_adaptive_results}\\
\toprule
\textbf{Benchmark} & \textbf{Method} & \textbf{MAE} $\downarrow$ & \textbf{Avg. Item} $\downarrow$ & \textbf{Test Overlap} $\downarrow$ & \textbf{Exposure Rate} $\downarrow$ & \textbf{Avg. Time (s)} $\downarrow$ \\
\midrule
\endhead

\toprule
\textbf{Benchmark} & \textbf{Method} & \textbf{MAE} $\downarrow$ & \textbf{Avg. Item} $\downarrow$ & \textbf{Test Overlap} $\downarrow$ & \textbf{Exposure Rate} $\downarrow$ & \textbf{Avg. Time (s)} $\downarrow$ \\
\midrule
\endhead

\midrule
\multicolumn{7}{r}{\textit{Continued on next page}}\\
\bottomrule
\endfoot

\bottomrule
\endlastfoot


\multirow{3}{*}{\makecell[c]{MMLU-Abstract Algebra\\(79 items)}} 
& ATLAS$_{SE \leq 0.1}$ & \textbf{0.025} & 52.45 & 0.6808 & 0.6639 & 0.59 \\
& ATLAS$_{SE \leq 0.2}$ & 0.067 & 32.18 & 0.4413 & 0.4073 & 0.36 \\
& ATLAS$_{SE \leq 0.3}$ & 0.098 & \textbf{19.90} & \textbf{0.3169} & \textbf{0.2519} & \textbf{0.22} \\
\midrule

\multirow{3}{*}{\makecell[c]{MMLU-Anatomy\\(113 items)}} 
& ATLAS$_{SE \leq 0.1}$ & \textbf{0.099} & 100.00 & 0.94 & 0.88 & 4.93 \\
& ATLAS$_{SE \leq 0.2}$ & 0.149 & 53.29 & 0.54 & 0.47 & 2.70 \\
& ATLAS$_{SE \leq 0.3}$ & 0.235 & \textbf{20.49} & \textbf{0.32} & \textbf{0.18} & \textbf{0.98} \\
\midrule

\multirow{3}{*}{\makecell[c]{MMLU-Astronomy\\(136 items)}} 
& ATLAS$_{SE \leq 0.1}$ & \textbf{0.099} & 93.22 & 0.82 & 0.69 & 6.30 \\
& ATLAS$_{SE \leq 0.2}$ & 0.157 & 48.21 & 0.48 & 0.35 & 3.29 \\
& ATLAS$_{SE \leq 0.3}$ & 0.235 & \textbf{11.80} & \textbf{0.40} & \textbf{0.11} & \textbf{0.79} \\
\midrule

\multirow{3}{*}{\makecell[c]{MMLU-Business Ethics\\(95 items)}} 
& ATLAS$_{SE \leq 0.1}$ & \textbf{0.040} & 86.61 & 0.9122 & 0.9117 & 4.09 \\
& ATLAS$_{SE \leq 0.2}$ & 0.072 & 54.43 & 0.5912 & 0.5729 & 5.29 \\
& ATLAS$_{SE \leq 0.3}$ & 0.160 & \textbf{13.46} & \textbf{0.4361} & \textbf{0.1417} & \textbf{0.63} \\
\midrule

\multirow{3}{*}{\makecell[c]{MMLU-Clinical Knowledge\\(198 items)}} 
& ATLAS$_{SE \leq 0.1}$ & \textbf{0.044} & 99.86 & 0.7302 & 0.5043 & 8.69 \\
& ATLAS$_{SE \leq 0.2}$ & 0.186 & 23.33 & 0.2894 & 0.1477 & 4.20 \\
& ATLAS$_{SE \leq 0.3}$ & 0.244 & \textbf{10.46} & \textbf{0.2537} & \textbf{0.1016} & \textbf{0.91} \\
\midrule

\multirow{3}{*}{\makecell[c]{MMLU-College Biology\\(131 items)}} 
& ATLAS$_{SE \leq 0.1}$ & \textbf{0.031} & 100.00 & 0.8870 & 0.7634 & 5.98 \\
& ATLAS$_{SE \leq 0.2}$ & 0.083 & 67.87 & 0.6168 & 0.5186 & 4.20 \\
& ATLAS$_{SE \leq 0.3}$ & 0.134 & \textbf{29.40} & \textbf{0.3380} & \textbf{0.2315} & \textbf{1.90} \\
\midrule

\multirow{3}{*}{\makecell[c]{MMLU-College Chemistry\\(77 items)}} 
& ATLAS$_{SE \leq 0.1}$ & \textbf{0.029} & 77.00 & 1.0000 & 1.0000 & 0.72 \\
& ATLAS$_{SE \leq 0.2}$ & 0.070 & 52.23 & 0.6857 & 0.6783 & 0.47 \\
& ATLAS$_{SE \leq 0.3}$ & 0.116 & \textbf{17.84} & \textbf{0.4230} & \textbf{0.2308} & \textbf{0.25} \\
\midrule

\multirow{3}{*}{\makecell[c]{MMLU-College Computer Science\\(84 items)}} 
& ATLAS$_{SE \leq 0.1}$ & \textbf{0.030} & 82.71 & 0.9845 & 0.9844 & 0.73 \\
& ATLAS$_{SE \leq 0.2}$ & 0.066 & 41.67 & 0.5305 & 0.4961 & 0.39 \\
& ATLAS$_{SE \leq 0.3}$ & 0.094 & \textbf{13.59} & \textbf{0.3367} & \textbf{0.1621} & \textbf{0.19} \\
\midrule

\multirow{3}{*}{\makecell[c]{MMLU-College Mathematics\\(69 items)}} 
& ATLAS$_{SE \leq 0.1}$ & \textbf{0.025} & 63.86 & 0.9267 & 0.9256 & 0.62 \\
& ATLAS$_{SE \leq 0.2}$ & 0.062 & 44.67 & 0.6669 & 0.6478 & 0.43 \\
& ATLAS$_{SE \leq 0.3}$ & 0.089 & \textbf{28.48} & \textbf{0.4832} & \textbf{0.4128} & \textbf{0.29} \\
\midrule

\multirow{3}{*}{\makecell[c]{MMLU-College Medicine\\(157 items)}} 
& ATLAS$_{SE \leq 0.1}$ & \textbf{0.038} & 100.00 & 0.8039 & 0.6369 & 7.02 \\
& ATLAS$_{SE \leq 0.2}$ & 0.111 & 66.64 & 0.5986 & 0.4350 & 4.90 \\
& ATLAS$_{SE \leq 0.3}$ & 0.171 & \textbf{27.77} & \textbf{0.3062} & \textbf{0.2061} & \textbf{2.02} \\
\midrule

\multirow{3}{*}{\makecell[c]{MMLU-College Physics\\(72 items)}} 
& ATLAS$_{SE \leq 0.1}$ & \textbf{0.020} & 69.39 & 0.9634 & 0.9632 & 0.69 \\
& ATLAS$_{SE \leq 0.2}$ & 0.041 & 61.78 & 0.8618 & 0.8580 & 0.63 \\
& ATLAS$_{SE \leq 0.3}$ & 0.074 & \textbf{30.69} & \textbf{0.5167} & \textbf{0.4264} & \textbf{0.36} \\
\midrule

\multirow{3}{*}{\makecell[c]{MMLU-Computer Security\\(84 items)}} 
& ATLAS$_{SE \leq 0.1}$ & \textbf{0.019} & 84.00 & 1.0000 & 1.0000 & 0.71 \\
& ATLAS$_{SE \leq 0.2}$ & 0.034 & 75.99 & 0.9067 & 0.9049 & 0.65 \\
& ATLAS$_{SE \leq 0.3}$ & 0.065 & \textbf{29.54} & \textbf{0.4484} & \textbf{0.3508} & \textbf{0.35} \\
\midrule

\multirow{3}{*}{\makecell[c]{MMLU-Conceptual Physics\\(203 items)}} 
& ATLAS$_{SE \leq 0.1}$ & \textbf{0.042} & 69.43 & 0.5217 & 0.3417 & 4.02 \\
& ATLAS$_{SE \leq 0.2}$ & 0.160 & 13.86 & 0.1844 & 0.0713 & 0.98 \\
& ATLAS$_{SE \leq 0.3}$ & 0.205 & \textbf{10.58} & \textbf{0.1844} & \textbf{0.0596} & \textbf{0.78} \\
\midrule

\multirow{3}{*}{\makecell[c]{MMLU-Econometrics\\(102 items)}} 
& ATLAS$_{SE \leq 0.1}$ & \textbf{0.031} & 91.67 & 0.9026 & 0.8992 & 5.38 \\
& ATLAS$_{SE \leq 0.2}$ & 0.074 & 56.23 & 0.5678 & 0.5508 & 4.05 \\
& ATLAS$_{SE \leq 0.3}$ & 0.148 & \textbf{13.44} & \textbf{0.4002} & \textbf{0.1442} & \textbf{0.64} \\
\midrule

\multirow{3}{*}{\makecell[c]{MMLU-Electrical Engineering\\(126 items)}} 
& ATLAS$_{SE \leq 0.1}$ & \textbf{0.032} & 100.00 & 0.8917 & 0.7937 & 5.98 \\
& ATLAS$_{SE \leq 0.2}$ & 0.081 & 61.94 & 0.5710 & 0.4917 & 4.05 \\
& ATLAS$_{SE \leq 0.3}$ & 0.134 & \textbf{18.70} & \textbf{0.3021} & \textbf{0.1498} & \textbf{1.05} \\
\midrule

\multirow{3}{*}{\makecell[c]{MMLU-Elementary Mathematics\\(220 items)}} 
& ATLAS$_{SE \leq 0.1}$ & \textbf{0.050} & 66.10 & 0.4546 & 0.3006 & 3.80 \\
& ATLAS$_{SE \leq 0.2}$ & 0.191 & 13.34 & 0.2313 & 0.0843 & 0.87 \\
& ATLAS$_{SE \leq 0.3}$ & 0.237 & \textbf{10.02} & \textbf{0.2020} & \textbf{0.0652} & \textbf{0.67} \\
\midrule

\multirow{3}{*}{\makecell[c]{MMLU-Formal Logic\\(109 items)}} 
& ATLAS$_{SE \leq 0.1}$ & \textbf{0.030} & 91.05 & 0.8751 & 0.8359 & 5.43 \\
& ATLAS$_{SE \leq 0.2}$ & 0.098 & 31.64 & 0.3807 & 0.2895 & 2.18 \\
& ATLAS$_{SE \leq 0.3}$ & 0.146 & \textbf{20.89} & \textbf{0.2762} & \textbf{0.1919} & \textbf{1.26} \\
\midrule

\multirow{3}{*}{\makecell[c]{MMLU-Global Facts\\(81 items)}} 
& ATLAS$_{SE \leq 0.1}$ & \textbf{0.022} & 65.09 & 0.8060 & 0.8041 & 0.56 \\
& ATLAS$_{SE \leq 0.2}$ & 0.038 & 57.62 & 0.7158 & 0.7115 & 0.50 \\
& ATLAS$_{SE \leq 0.3}$ & 0.069 & \textbf{18.03} & \textbf{0.4382} & \textbf{0.2225} & \textbf{0.22} \\
\midrule

\multirow{3}{*}{\makecell[c]{MMLU-High School Biology\\(251 items)}} 
& ATLAS$_{SE \leq 0.1}$ & \textbf{0.037} & 100.00 & 0.6336 & 0.3984 & 6.09 \\
& ATLAS$_{SE \leq 0.2}$ & 0.124 & 37.74 & 0.2869 & 0.1932 & 2.78 \\
& ATLAS$_{SE \leq 0.3}$ & 0.171 & \textbf{25.49} & \textbf{0.2193} & \textbf{0.1480} & \textbf{1.98} \\
\midrule

\multirow{3}{*}{\makecell[c]{MMLU-High School Chemistry\\(169 items)}} 
& ATLAS$_{SE \leq 0.1}$ & \textbf{0.033} & 100.00 & 0.7729 & 0.5917 & 5.64 \\
& ATLAS$_{SE \leq 0.2}$ & 0.111 & 36.56 & 0.3257 & 0.2349 & 2.62 \\
& ATLAS$_{SE \leq 0.3}$ & 0.151 & \textbf{17.30} & \textbf{0.2078} & \textbf{0.1182} & \textbf{1.00} \\
\midrule

\multirow{3}{*}{\makecell[c]{MMLU-High School Computer Science\\(94 items)}} 
& ATLAS$_{SE \leq 0.1}$ & \textbf{0.021} & 94.00 & 1.0000 & 1.0000 & 0.74 \\
& ATLAS$_{SE \leq 0.2}$ & 0.041 & 73.23 & 0.7911 & 0.7788 & 0.59 \\
& ATLAS$_{SE \leq 0.3}$ & 0.066 & \textbf{31.34} & \textbf{0.4380} & \textbf{0.3330} & \textbf{0.34} \\
\midrule

\multirow{3}{*}{\makecell[c]{MMLU-High School European History\\(147 items)}} 
& ATLAS$_{SE \leq 0.1}$ & \textbf{0.038} & 100.00 & 0.8630 & 0.6803 & 6.31 \\
& ATLAS$_{SE \leq 0.2}$ & 0.081 & 75.10 & 0.6755 & 0.5368 & 5.16 \\
& ATLAS$_{SE \leq 0.3}$ & 0.137 & \textbf{27.05} & \textbf{0.3238} & \textbf{0.2001} & \textbf{2.15} \\
\midrule

\multirow{3}{*}{\makecell[c]{MMLU-High School Geography\\(170 items)}} 
& ATLAS$_{SE \leq 0.1}$ & \textbf{0.036} & 85.66 & 0.6861 & 0.5039 & 4.89 \\
& ATLAS$_{SE \leq 0.2}$ & 0.095 & 45.02 & 0.4002 & 0.2904 & 2.90 \\
& ATLAS$_{SE \leq 0.3}$ & 0.142 & \textbf{17.80} & \textbf{0.3141} & \textbf{0.1356} & \textbf{1.19} \\
\midrule

\multirow{3}{*}{\makecell[c]{MMLU-High School Government \& Politics\\(154 items)}} 
& ATLAS$_{SE \leq 0.1}$ & \textbf{0.033} & 100.00 & 0.8509 & 0.6494 & 5.94 \\
& ATLAS$_{SE \leq 0.2}$ & 0.062 & 92.63 & 0.7887 & 0.6011 & 5.54 \\
& ATLAS$_{SE \leq 0.3}$ & 0.124 & \textbf{26.66} & \textbf{0.3425} & \textbf{0.1989} & \textbf{1.69} \\
\midrule

\multirow{3}{*}{\makecell[c]{MMLU-High School Mathematics\\(199 items)}} 
& ATLAS$_{SE \leq 0.1}$ & \textbf{0.044} & 56.19 & 0.4137 & 0.2822 & 3.31 \\
& ATLAS$_{SE \leq 0.2}$ & 0.094 & 47.18 & 0.3732 & 0.2486 & 2.86 \\
& ATLAS$_{SE \leq 0.3}$ & 0.183 & \textbf{12.12} & \textbf{0.2503} & \textbf{0.0894} & \textbf{0.79} \\
\midrule

\multirow{3}{*}{\makecell[c]{MMLU-High School Microeconomics\\(193 items)}} 
& ATLAS$_{SE \leq 0.1}$ & \textbf{0.040} & 93.17 & 0.7021 & 0.4828 & 5.55 \\
& ATLAS$_{SE \leq 0.2}$ & 0.174 & 18.27 & 0.2907 & 0.1188 & 1.25 \\
& ATLAS$_{SE \leq 0.3}$ & 0.223 & \textbf{10.64} & \textbf{0.2579} & \textbf{0.1005} & \textbf{0.79} \\
\midrule

\multirow{3}{*}{\makecell[c]{MMLU-High School\\Macroeconomics (281 items)}} 
& ATLAS$_{SE \leq 0.1}$ & \textbf{0.089} & 87.94 & 0.48 & 0.31 & 10.49 \\
& ATLAS$_{SE \leq 0.2}$ & 0.197 & 27.64 & 0.22 & 0.13 & 3.46 \\
& ATLAS$_{SE \leq 0.3}$ & 0.240 & \textbf{18.26} & \textbf{0.20} & \textbf{0.09} & \textbf{2.34} \\
\midrule

\multirow{3}{*}{\makecell[c]{MMLU-High School Physics\\(119 items)}} 
& ATLAS$_{SE \leq 0.1}$ & \textbf{0.026} & 63.30 & 0.5825 & 0.5317 & 3.54 \\
& ATLAS$_{SE \leq 0.2}$ & 0.055 & 51.75 & 0.4900 & 0.4356 & 3.08 \\
& ATLAS$_{SE \leq 0.3}$ & 0.106 & \textbf{20.30} & \textbf{0.3393} & \textbf{0.1932} & \textbf{1.37} \\
\midrule

\multirow{3}{*}{\makecell[c]{MMLU-High School Psychology\\(285 items)}} 
& ATLAS$_{SE \leq 0.1}$ & \textbf{0.035} & 100.00 & 0.6086 & 0.3509 & 6.19 \\
& ATLAS$_{SE \leq 0.2}$ & 0.147 & 26.80 & 0.2433 & 0.1284 & 1.97 \\
& ATLAS$_{SE \leq 0.3}$ & 0.196 & \textbf{11.48} & \textbf{0.2101} & \textbf{0.0881} & \textbf{0.92} \\
\midrule

\multirow{3}{*}{\makecell[c]{MMLU-High School Statistics\\(185 items)}} 
& ATLAS$_{SE \leq 0.1}$ & \textbf{0.038} & 69.47 & 0.5587 & 0.3757 & 4.07 \\
& ATLAS$_{SE \leq 0.2}$ & 0.154 & 21.48 & 0.2987 & 0.1219 & 1.67 \\
& ATLAS$_{SE \leq 0.3}$ & 0.205 & \textbf{11.81} & \textbf{0.2483} & \textbf{0.0700} & \textbf{1.03} \\
\midrule

\multirow{3}{*}{\makecell[c]{MMLU-High School US History\\(183 items)}} 
& ATLAS$_{SE \leq 0.1}$ & \textbf{0.030} & 96.68 & 0.7761 & 0.5283 & 5.77 \\
& ATLAS$_{SE \leq 0.2}$ & 0.069 & 73.31 & 0.6137 & 0.4027 & 4.80 \\
& ATLAS$_{SE \leq 0.3}$ & 0.151 & \textbf{13.62} & \textbf{0.3348} & \textbf{0.1319} & \textbf{1.05} \\
\midrule

\multirow{3}{*}{\makecell[c]{MMLU-High School World History\\(206 items)}} 
& ATLAS$_{SE \leq 0.1}$ & \textbf{0.031} & 96.43 & 0.7407 & 0.4680 & 5.88 \\
& ATLAS$_{SE \leq 0.2}$ & 0.108 & 40.83 & 0.3806 & 0.2580 & 2.85 \\
& ATLAS$_{SE \leq 0.3}$ & 0.176 & \textbf{17.10} & \textbf{0.2629} & \textbf{0.1230} & \textbf{1.28} \\
\midrule

\multirow{3}{*}{\makecell[c]{MMLU-Human Aging\\(188 items)}} 
& ATLAS$_{SE \leq 0.1}$ & \textbf{0.034} & 100.00 & 0.7341 & 0.5319 & 5.86 \\
& ATLAS$_{SE \leq 0.2}$ & 0.108 & 40.24 & 0.3546 & 0.2515 & 2.69 \\
& ATLAS$_{SE \leq 0.3}$ & 0.154 & \textbf{27.38} & \textbf{0.2445} & \textbf{0.1802} & \textbf{2.00} \\
\midrule

\multirow{3}{*}{\makecell[c]{MMLU-Human Sexuality\\(116 items)}} 
& ATLAS$_{SE \leq 0.1}$ & \textbf{0.032} & 100.00 & 0.9137 & 0.8621 & 6.07 \\
& ATLAS$_{SE \leq 0.2}$ & 0.062 & 83.66 & 0.7655 & 0.7212 & 5.25 \\
& ATLAS$_{SE \leq 0.3}$ & 0.154 & \textbf{20.43} & \textbf{0.2825} & \textbf{0.1759} & \textbf{1.51} \\
\midrule

\multirow{3}{*}{\makecell[c]{MMLU-International Law\\(103 items)}} 
& ATLAS$_{SE \leq 0.1}$ & \textbf{0.028} & 100.00 & 0.9780 & 0.9709 & 5.88 \\
& ATLAS$_{SE \leq 0.2}$ & 0.061 & 76.17 & 0.7582 & 0.7400 & 4.72 \\
& ATLAS$_{SE \leq 0.3}$ & 0.105 & \textbf{34.28} & \textbf{0.4298} & \textbf{0.3333} & \textbf{2.57} \\
\midrule

\multirow{3}{*}{\makecell[c]{MMLU-Jurisprudence\\(99 items)}} 
& ATLAS$_{SE \leq 0.1}$ & \textbf{0.021} & 99.00 & 1.0000 & 1.0000 & 0.79 \\
& ATLAS$_{SE \leq 0.2}$ & 0.047 & 71.20 & 0.7262 & 0.7191 & 0.60 \\
& ATLAS$_{SE \leq 0.3}$ & 0.095 & \textbf{33.57} & \textbf{0.3986} & \textbf{0.3399} & \textbf{0.38} \\
\midrule

\multirow{3}{*}{\makecell[c]{MMLU-Logical Fallacies\\(147 items)}} 
& ATLAS$_{SE \leq 0.1}$ & \textbf{0.030} & 100.00 & 0.8562 & 0.6803 & 5.92 \\
& ATLAS$_{SE \leq 0.2}$ & 0.064 & 70.48 & 0.6483 & 0.4928 & 4.73 \\
& ATLAS$_{SE \leq 0.3}$ & 0.117 & \textbf{22.21} & \textbf{0.2938} & \textbf{0.1632} & \textbf{1.69} \\
\midrule

\multirow{3}{*}{\makecell[c]{MMLU-Machine Learning\\(98 items)}} 
& ATLAS$_{SE \leq 0.1}$ & \textbf{0.027} & 74.29 & 0.7619 & 0.7586 & 3.34 \\
& ATLAS$_{SE \leq 0.2}$ & 0.069 & 39.44 & 0.4402 & 0.4018 & 1.89 \\
& ATLAS$_{SE \leq 0.3}$ & 0.143 & \textbf{13.81} & \textbf{0.3032} & \textbf{0.1410} & \textbf{0.71} \\
\midrule

\multirow{3}{*}{\makecell[c]{MMLU-Management\\(92 items)}} 
& ATLAS$_{SE \leq 0.1}$ & \textbf{0.022} & 92.00 & 1.0000 & 1.0000 & 0.74 \\
& ATLAS$_{SE \leq 0.2}$ & 0.024 & 90.39 & 0.9831 & 0.9831 & 0.72 \\
& ATLAS$_{SE \leq 0.3}$ & 0.059 & \textbf{44.91} & \textbf{0.5238} & \textbf{0.4881} & \textbf{0.47} \\
\midrule

\multirow{3}{*}{\makecell[c]{MMLU-Marketing\\(178 items)}} 
& ATLAS$_{SE \leq 0.1}$ & \textbf{0.033} & 100.00 & 0.7974 & 0.5618 & 5.78 \\
& ATLAS$_{SE \leq 0.2}$ & 0.081 & 61.18 & 0.5305 & 0.3645 & 4.06 \\
& ATLAS$_{SE \leq 0.3}$ & 0.134 & \textbf{25.09} & \textbf{0.3035} & \textbf{0.1660} & \textbf{1.84} \\
\midrule

\multirow{3}{*}{\makecell[c]{MMLU-Medical Genetics\\(87 items)}} 
& ATLAS$_{SE \leq 0.1}$ & \textbf{0.024} & 84.38 & 0.9699 & 0.9698 & 0.74 \\
& ATLAS$_{SE \leq 0.2}$ & 0.045 & 69.31 & 0.7995 & 0.7967 & 0.62 \\
& ATLAS$_{SE \leq 0.3}$ & 0.072 & \textbf{55.16} & \textbf{0.6449} & \textbf{0.6349} & \textbf{0.51} \\
\midrule

\multirow{3}{*}{\makecell[c]{MMLU-Miscellaneous\\(314 items)}} 
& ATLAS$_{SE \leq 0.1}$ & \textbf{0.048} & 65.51 & 0.3154 & 0.2085 & 3.66 \\
& ATLAS$_{SE \leq 0.2}$ & 0.156 & 24.00 & 0.1576 & 0.0991 & 1.58 \\
& ATLAS$_{SE \leq 0.3}$ & 0.203 & \textbf{15.73} & \textbf{0.1295} & \textbf{0.0697} & \textbf{1.10} \\
\midrule

\multirow{3}{*}{\makecell[c]{MMLU-Moral Disputes\\(229 items)}} 
& ATLAS$_{SE \leq 0.1}$ & \textbf{0.041} & 99.73 & 0.6574 & 0.4355 & 6.03 \\
& ATLAS$_{SE \leq 0.2}$ & 0.142 & 27.23 & 0.2498 & 0.1526 & 1.94 \\
& ATLAS$_{SE \leq 0.3}$ & 0.195 & \textbf{13.45} & \textbf{0.2077} & \textbf{0.1090} & \textbf{0.98} \\
\midrule

\multirow{3}{*}{\makecell[c]{MMLU-Moral Scenarios\\(486 items)}} 
& ATLAS$_{SE \leq 0.1}$ & \textbf{0.059} & 34.18 & 0.1610 & 0.0704 & 2.24 \\
& ATLAS$_{SE \leq 0.2}$ & 0.114 & 19.23 & 0.1216 & 0.0436 & 1.41 \\
& ATLAS$_{SE \leq 0.3}$ & 0.158 & \textbf{14.14} & \textbf{0.0986} & \textbf{0.0334} & \textbf{1.05} \\
\midrule

\multirow{3}{*}{\makecell[c]{MMLU-Nutrition\\(244 items)}} 
& ATLAS$_{SE \leq 0.1}$ & \textbf{0.039} & 95.50 & 0.6325 & 0.3914 & 6.00 \\
& ATLAS$_{SE \leq 0.2}$ & 0.129 & 38.88 & 0.3028 & 0.1953 & 2.89 \\
& ATLAS$_{SE \leq 0.3}$ & 0.176 & \textbf{21.25} & \textbf{0.2221} & \textbf{0.1191} & \textbf{1.69} \\
\midrule

\multirow{3}{*}{\makecell[c]{MMLU-Philosophy\\(214 items)}} 
& ATLAS$_{SE \leq 0.1}$ & \textbf{0.041} & 96.45 & 0.6360 & 0.4510 & 5.90 \\
& ATLAS$_{SE \leq 0.2}$ & 0.152 & 23.09 & 0.1983 & 0.1258 & 1.89 \\
& ATLAS$_{SE \leq 0.3}$ & 0.202 & \textbf{16.56} & \textbf{0.1657} & \textbf{0.0927} & \textbf{1.37} \\
\midrule

\multirow{3}{*}{\makecell[c]{MMLU-Prehistory\\(265 items)}} 
& ATLAS$_{SE \leq 0.1}$ & \textbf{0.039} & 93.64 & 0.6004 & 0.3531 & 5.72 \\
& ATLAS$_{SE \leq 0.2}$ & 0.099 & 44.74 & 0.4126 & 0.2078 & 3.28 \\
& ATLAS$_{SE \leq 0.3}$ & 0.190 & \textbf{12.29} & \textbf{0.2773} & \textbf{0.0989} & \textbf{0.95} \\
\midrule

\multirow{3}{*}{\makecell[c]{MMLU-Professional Accounting\\(221 items)}} 
& ATLAS$_{SE \leq 0.1}$ & \textbf{0.047} & 63.18 & 0.4458 & 0.2861 & 3.83 \\
& ATLAS$_{SE \leq 0.2}$ & 0.170 & 14.16 & 0.2016 & 0.0682 & 1.03 \\
& ATLAS$_{SE \leq 0.3}$ & 0.220 & \textbf{10.46} & \textbf{0.1899} & \textbf{0.0647} & \textbf{0.83} \\
\midrule

\multirow{3}{*}{\makecell[c]{MMLU-Professional Law\\(518 items)}} 
& ATLAS$_{SE \leq 0.1}$ & \textbf{0.064} & 26.94 & 0.1811 & 0.0519 & 1.84 \\
& ATLAS$_{SE \leq 0.2}$ & 0.151 & 10.88 & 0.1380 & 0.0283 & 0.98 \\
& ATLAS$_{SE \leq 0.3}$ & 0.188 & \textbf{10.06} & \textbf{0.1308} & \textbf{0.0271} & \textbf{0.90} \\
\midrule

\multirow{3}{*}{\makecell[c]{MMLU-Professional Medicine\\(218 items)}} 
& ATLAS$_{SE \leq 0.1}$ & \textbf{0.044} & 95.14 & 0.6919 & 0.4363 & 6.11 \\
& ATLAS$_{SE \leq 0.2}$ & 0.167 & 24.91 & 0.3403 & 0.1649 & 2.12 \\
& ATLAS$_{SE \leq 0.3}$ & 0.220 & \textbf{10.25} & \textbf{0.2794} & \textbf{0.1019} & \textbf{0.95} \\
\midrule

\multirow{3}{*}{\makecell[c]{MMLU-Professional Psychology\\(344 items)}} 
& ATLAS$_{SE \leq 0.1}$ & \textbf{0.051} & 58.00 & 0.3224 & 0.1685 & 3.86 \\
& ATLAS$_{SE \leq 0.2}$ & 0.166 & 11.53 & 0.1850 & 0.0670 & 1.06 \\
& ATLAS$_{SE \leq 0.3}$ & 0.205 & \textbf{10.08} & \textbf{0.1715} & \textbf{0.0645} & \textbf{0.95} \\
\midrule

\multirow{3}{*}{\makecell[c]{MMLU-Public Relations\\(93 items)}} 
& ATLAS$_{SE \leq 0.1}$ & \textbf{0.027} & 93.05 & 1.0000 & 1.0000 & 0.74 \\
& ATLAS$_{SE \leq 0.2}$ & 0.048 & 72.97 & 0.7926 & 0.7854 & 0.61 \\
& ATLAS$_{SE \leq 0.3}$ & 0.095 & \textbf{24.50} & \textbf{0.3729} & \textbf{0.2633} & \textbf{0.31} \\
\midrule

\multirow{3}{*}{\makecell[c]{MMLU-Security Studies\\(201 items)}} 
& ATLAS$_{SE \leq 0.1}$ & \textbf{0.038} & 100.00 & 0.7247 & 0.4975 & 6.11 \\
& ATLAS$_{SE \leq 0.2}$ & 0.119 & 34.74 & 0.3103 & 0.2038 & 2.59 \\
& ATLAS$_{SE \leq 0.3}$ & 0.171 & \textbf{17.12} & \textbf{0.1992} & \textbf{0.1103} & \textbf{1.30} \\
\midrule

\multirow{3}{*}{\makecell[c]{MMLU-Sociology\\(166 items)}} 
& ATLAS$_{SE \leq 0.1}$ & \textbf{0.032} & 100.00 & 0.8412 & 0.6024 & 6.02 \\
& ATLAS$_{SE \leq 0.2}$ & 0.070 & 72.92 & 0.6391 & 0.5095 & 5.00 \\
& ATLAS$_{SE \leq 0.3}$ & 0.137 & \textbf{37.22} & \textbf{0.3946} & \textbf{0.2712} & \textbf{2.70} \\
\midrule

\multirow{3}{*}{\makecell[c]{MMLU-US Foreign Policy\\(90 items)}} 
& ATLAS$_{SE \leq 0.1}$ & \textbf{0.024} & 90.00 & 1.0000 & 1.0000 & 0.72 \\
& ATLAS$_{SE \leq 0.2}$ & 0.026 & 89.55 & 0.9950 & 0.9950 & 0.71 \\
& ATLAS$_{SE \leq 0.3}$ & 0.063 & \textbf{30.48} & \textbf{0.4204} & \textbf{0.3389} & \textbf{0.37} \\
\midrule

\multirow{3}{*}{\makecell[c]{MMLU-Virology\\(140 items)}} 
& ATLAS$_{SE \leq 0.1}$ & \textbf{0.032} & 100.00 & 0.8644 & 0.7143 & 6.01 \\
& ATLAS$_{SE \leq 0.2}$ & 0.081 & 63.64 & 0.5787 & 0.4711 & 4.40 \\
& ATLAS$_{SE \leq 0.3}$ & 0.134 & \textbf{24.11} & \textbf{0.2965} & \textbf{0.1926} & \textbf{1.84} \\
\midrule

\multirow{3}{*}{\makecell[c]{MMLU-World Religions\\(137 items)}} 
& ATLAS$_{SE \leq 0.1}$ & \textbf{0.033} & 100.00 & 0.8796 & 0.7299 & 6.05 \\
& ATLAS$_{SE \leq 0.2}$ & 0.068 & 87.19 & 0.7734 & 0.6657 & 5.23 \\
& ATLAS$_{SE \leq 0.3}$ & 0.134 & \textbf{37.56} & \textbf{0.4063} & \textbf{0.2984} & \textbf{2.68} \\

\bottomrule
\label{tab:mmlu_adaptive_results}
\end{longtable}
\end{scriptsize}

\begin{figure}[ht]
    \centering
    \begin{subfigure}{\textwidth}
        \centering
TruthfulQA\\
\includegraphics[width=0.59\linewidth]{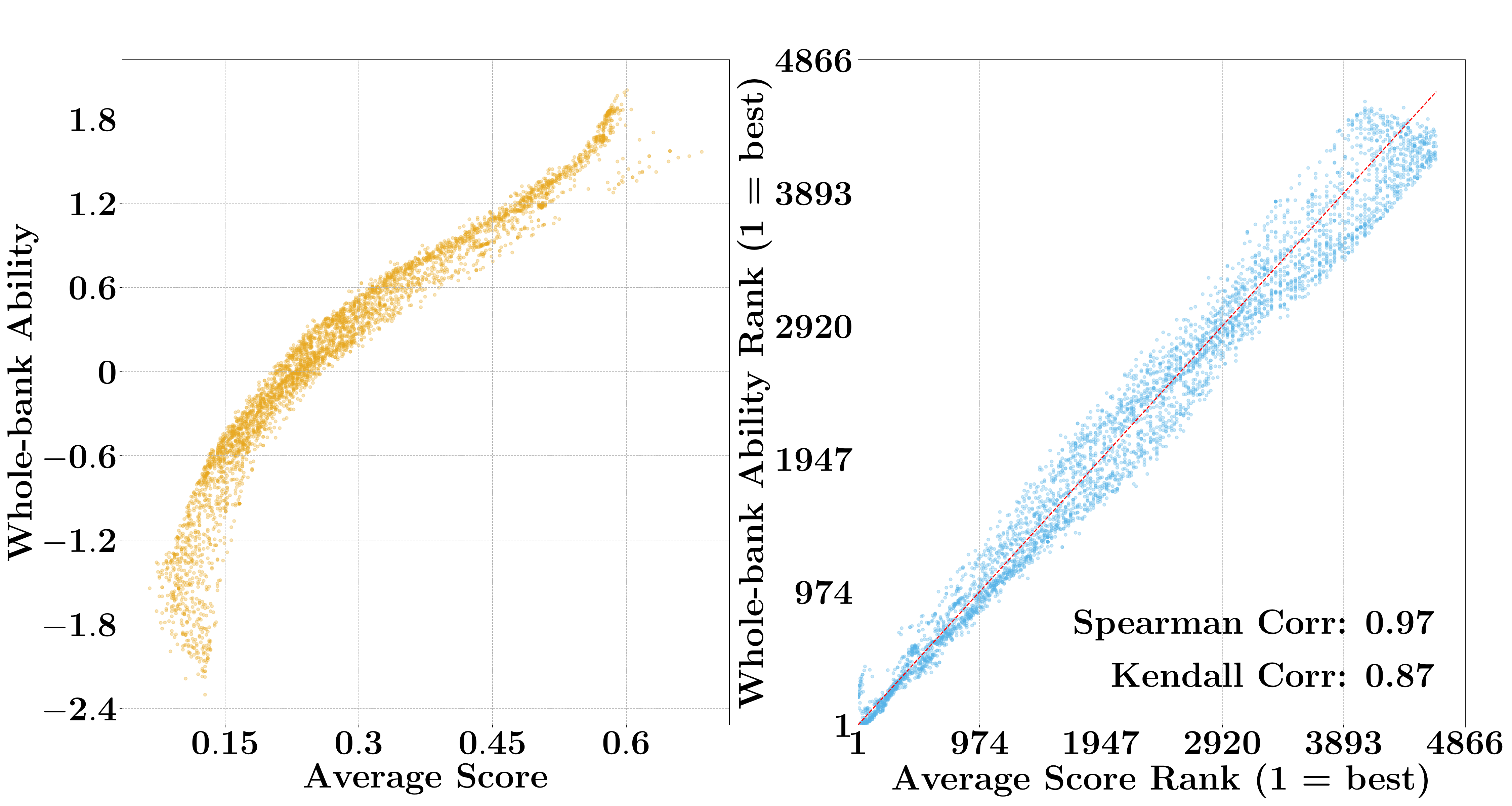}

\end{subfigure}
\caption{\textbf{Comparison of raw average scores and whole-bank ability estimates on TruthfulQA.} 
(Left) While average scores compress performance at the extremes, whole-bank ability estimates reveal clearer separation among both low- and high-performing models, reflecting sensitivity to item difficulty and discrimination. 
(Right) Rank comparison shows strong consistency between the two measures (Spearman $\rho = 0.97$, Kendall $\tau = 0.87$), but ability-based ranking provides finer resolution, especially in distinguishing weaker and stronger models beyond what raw accuracy captures.}

    \begin{subfigure}{\textwidth}
        \centering
WinoGrande\\
    \includegraphics[width=0.59\linewidth]{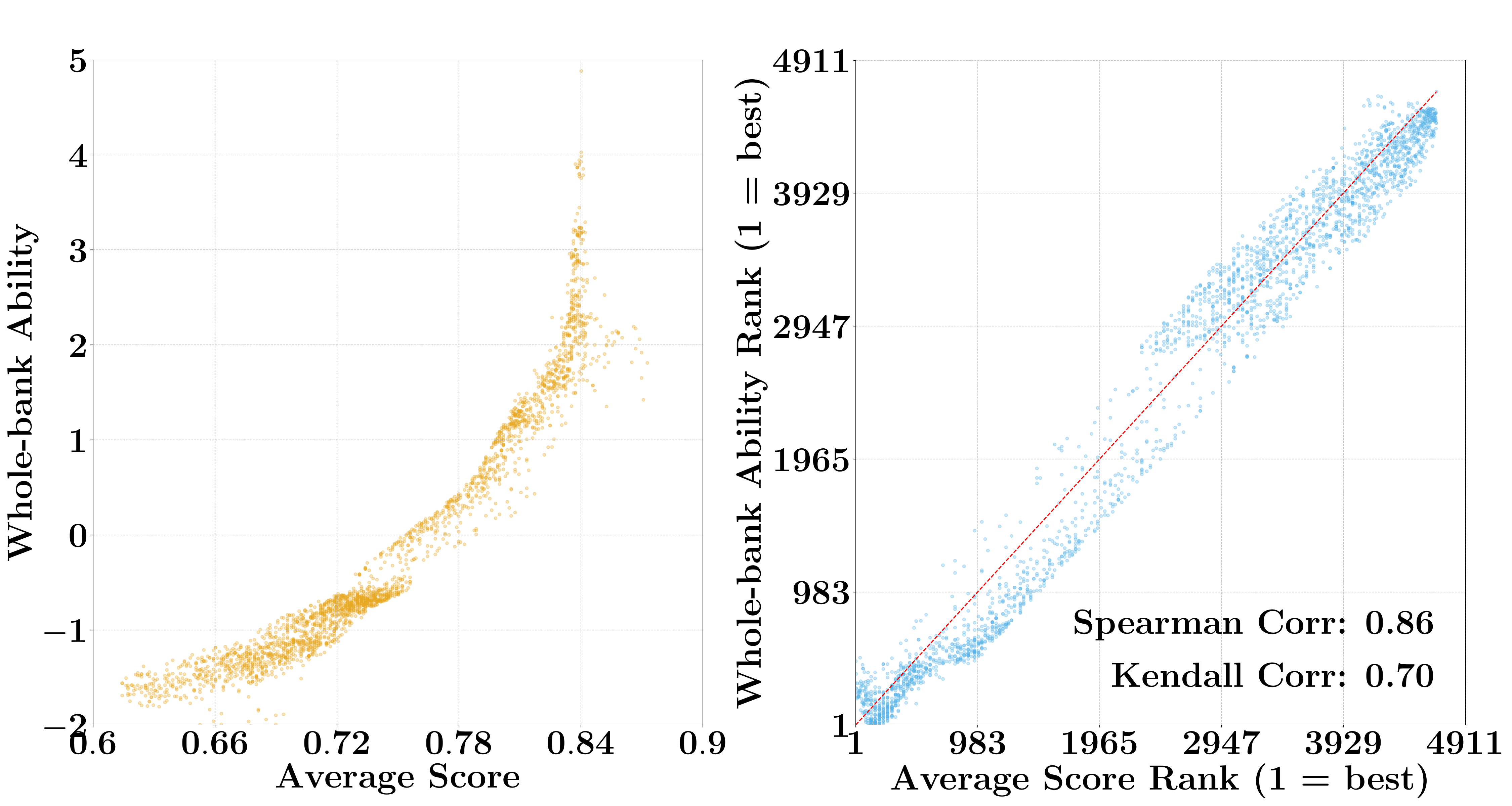}
    \end{subfigure}
   \caption{\textbf{Comparison of raw average scores and whole-bank ability estimates on WinoGrande.} 
(Left) Whole-bank ability estimates show a non-linear relationship with average score and reveal clearer separation on high-performing models, highlighting that ability captures relative item difficulty and provides finer differentiation beyond raw accuracy. 
(Right) Rank comparison indicates strong but imperfect alignment (Spearman $\rho = 0.86$, Kendall $\tau = 0.70$), with deviations from the diagonal reflecting cases where ability-based ranking distinguishes models more effectively than accuracy alone.}

    \begin{subfigure}{\textwidth}
        \centering
ARC\\
    \includegraphics[width=0.59\linewidth]{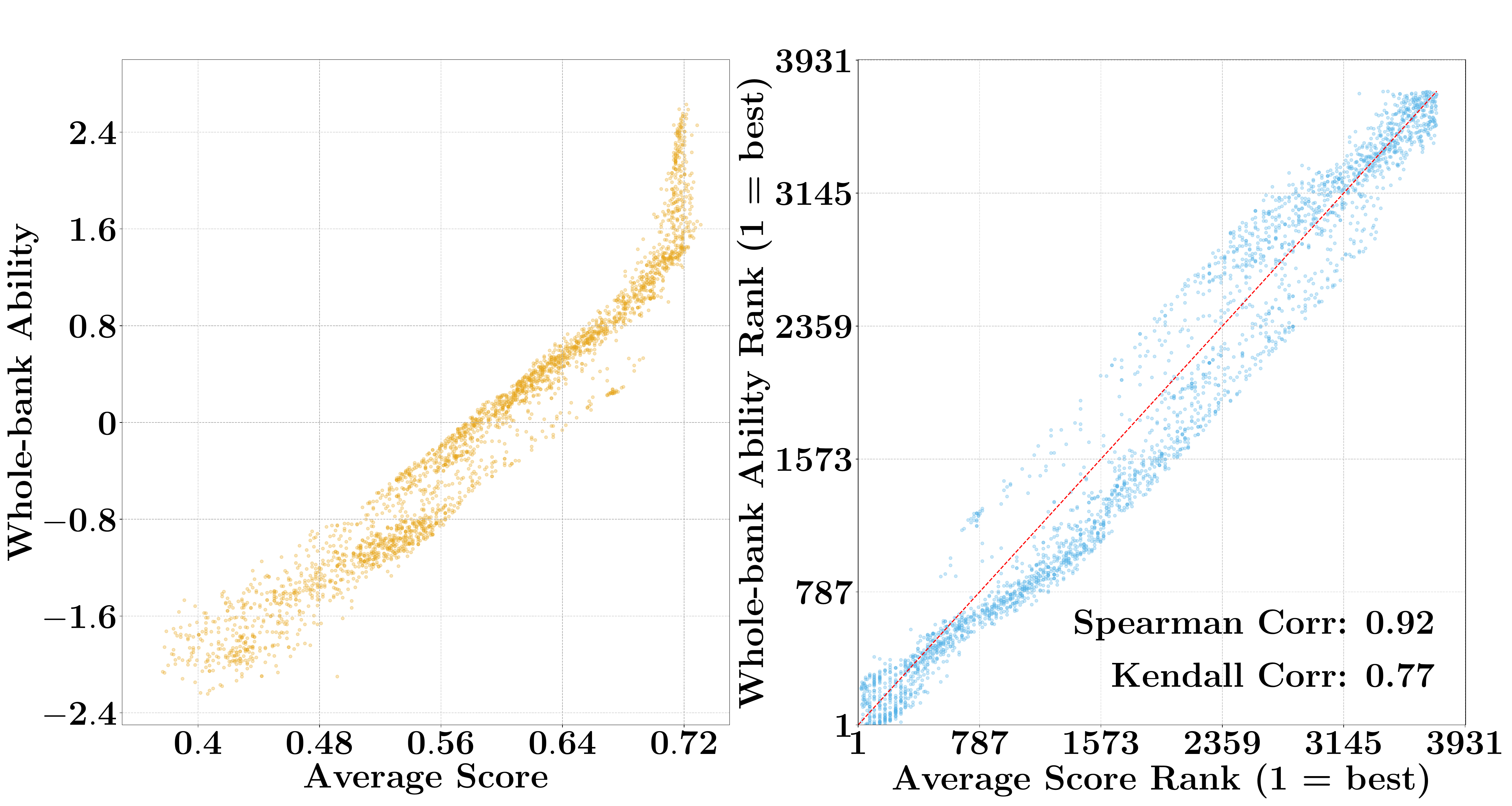}
        \end{subfigure}

   \caption{\textbf{Comparison of raw average scores and whole-bank ability estimates on ARC.} 
(Left) Whole-bank ability estimates exhibit a non-linear relationship with average scores, providing clearer separation on high-performing models by accounting for item difficulty and discrimination. 
(Right) Rank comparison shows strong but not perfect alignment between the two metrics (Spearman $\rho = 0.91$, Kendall $\tau = 0.77$), with deviations from the diagonal highlighting cases where ability-based ranking offers more informative distinctions than raw accuracy alone.}

\end{figure}

\begin{figure}[ht]
    \centering
    ARC\\
    \includegraphics[width=0.9\linewidth]{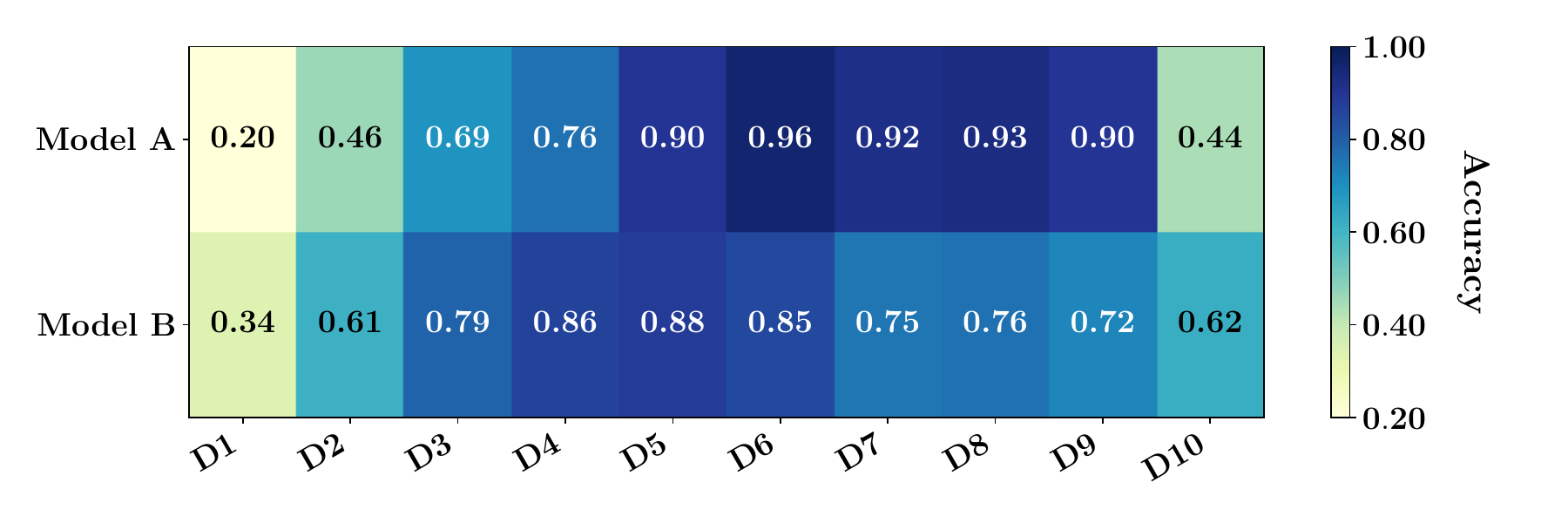}
    \caption{Two models with the similar average accuracy (0.713) and (0.714) on ARC nevertheless receive very different whole-bank ability estimates. Difficulty levels are ordered from easiest (D1) to hardest (D10). Model A (\texttt{mera-mix-4x7B}) attains a whole-bank ability rank of 270 because its correct responses are concentrated on more difficult items. In contrast, Model B (\texttt{LLaMAAntino-3-ANITA-8B-Inst-DPO-ITA}) is assigned a much lower whole-bank ability rank of 2612, as its successes occur primarily on easier items. This divergence shows how IRT-based ability estimation can distinguish models that appear identical under raw accuracy by accounting for item difficulty.
}
\end{figure}

\begin{figure}[ht]
    \centering
    HellaSwag\\   \includegraphics[width=0.85\linewidth]{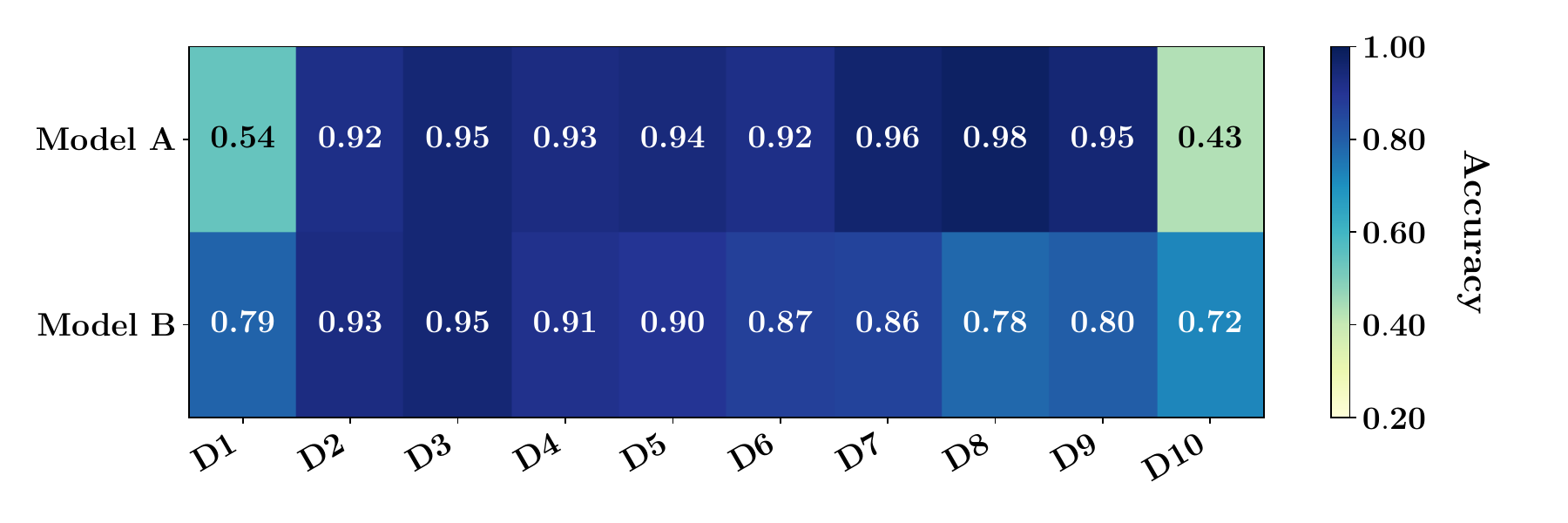}
    \caption{Two models with the same average accuracy (0.853) on HellaSwag nevertheless receive very different whole-bank ability estimates. Difficulty levels are ordered from easiest (D1) to hardest (D10). Model A (\texttt{supermario\_v1}) attains a whole-bank ability rank of 347 because its correct responses are concentrated on more difficult items. In contrast, Model B (\texttt{contaminated\_proof\_7b\_v1.0\_safetensor}) is assigned a much lower whole-bank ability rank of 3074, as its successes occur primarily on easier items. This divergence shows how IRT-based ability estimation can distinguish models that appear identical under raw accuracy by accounting for item difficulty.
}
\end{figure}

\end{document}